\documentclass[10pt,twocolumn,letterpaper]{article}

\usepackage{cvpr}              

\usepackage{graphicx}
\usepackage{amsmath}
\usepackage{amssymb}
\usepackage{booktabs}
\usepackage{comment}
\usepackage{xcolor}
\usepackage{multirow}
\usepackage{shortbold}
\usepackage{verbatim}
\usepackage{multirow}
\usepackage{caption}
\usepackage{array}
\usepackage{makecell}
\usepackage{verbatim}
\usepackage{bm}
\usepackage{colortbl}

\def\method{ViewSeg\xspace}
\def\cloud{CloudSeg\xspace}
\definecolor{AccessibleBlue}{rgb}{0.10196, 0.52157, 1.0}
\definecolor{AccessibleRed}{rgb}{0.8314, 0.0667, 0.3490}
\definecolor{lblue}{rgb}{0.9411, 0.9725, 1.0}
\definecolor{llblue}{rgb}{0.3, 0.6, 1.0}
\definecolor{lgreen}{rgb}{0.908, 0.961, 0.908}
\definecolor{llgreen}{rgb}{0.4, 0.7, 0.4}
\newcommand{\mypar}[1]{\vspace{1mm}\noindent\textbf{#1}}
\newcolumntype{L}[1]{>{\raggedright\arraybackslash}p{#1}}
\newcolumntype{C}[1]{>{\centering\arraybackslash}p{#1}}
\def\hhline{\Xhline{2\arrayrulewidth}}

\newcolumntype{S}[1]{>{\columncolor{lblue}\centering\arraybackslash}p{#1}}
\newcolumntype{D}[1]{>{\columncolor{lgreen}\centering\arraybackslash}p{#1}}

\usepackage[pagebackref,breaklinks,colorlinks]{hyperref}

\usepackage[capitalize]{cleveref}
\crefname{section}{Sec.}{Secs.}
\Crefname{section}{Section}{Sections}
\Crefname{table}{Table}{Tables}
\crefname{table}{Tab.}{Tabs.}

\def\cvprPaperID{3099} 
\def\confName{CVPR}
\def\confYear{2022}

\begin{document}

\title{Recognizing Scenes from Novel Viewpoints}


\author{
  Shengyi Qian$^\ddagger$ \hspace{2mm}
  Alexander Kirillov$^\dagger$ \hspace{2mm}
  Nikhila Ravi$^\dagger$ \hspace{2mm}
  Devendra Singh Chaplot$^\dagger$ \\*[2mm]
  Justin Johnson$^{\ddagger, \dagger}$ \hspace{2mm}
  David F. Fouhey$^\ddagger$ \hspace{2mm}
  Georgia Gkioxari$^\dagger$ \\*[3mm]
  $^\dagger$ Facebook AI Research \hspace{18mm} $^\ddagger$ University of Michigan, Ann Arbor
}

\maketitle

\begin{abstract}

Humans can perceive scenes in 3D from a handful of 2D views.
For AI agents, the ability to recognize a scene from any viewpoint given only a few images enables them to efficiently interact with the scene and its objects.
In this work, we attempt to endow machines with this ability.
We propose a model which takes as input a few RGB images of a new scene and recognizes the scene from novel viewpoints by segmenting it into semantic categories.
All this without access to the RGB images from those views.
We pair 2D scene recognition with an implicit 3D representation and learn from multi-view 2D annotations of hundreds of scenes without any 3D supervision beyond camera poses.
We experiment on challenging datasets and demonstrate our model's ability to jointly capture semantics and geometry of novel scenes with diverse layouts, object types and shapes.
\footnote{Project page: \url{https://jasonqsy.github.io/viewseg}.}
\end{abstract}

\vspace{-2mm}
\section{Introduction}
\label{sec:intro}

\begin{figure}[t]
  \centering
   \includegraphics[width=1.0\linewidth]{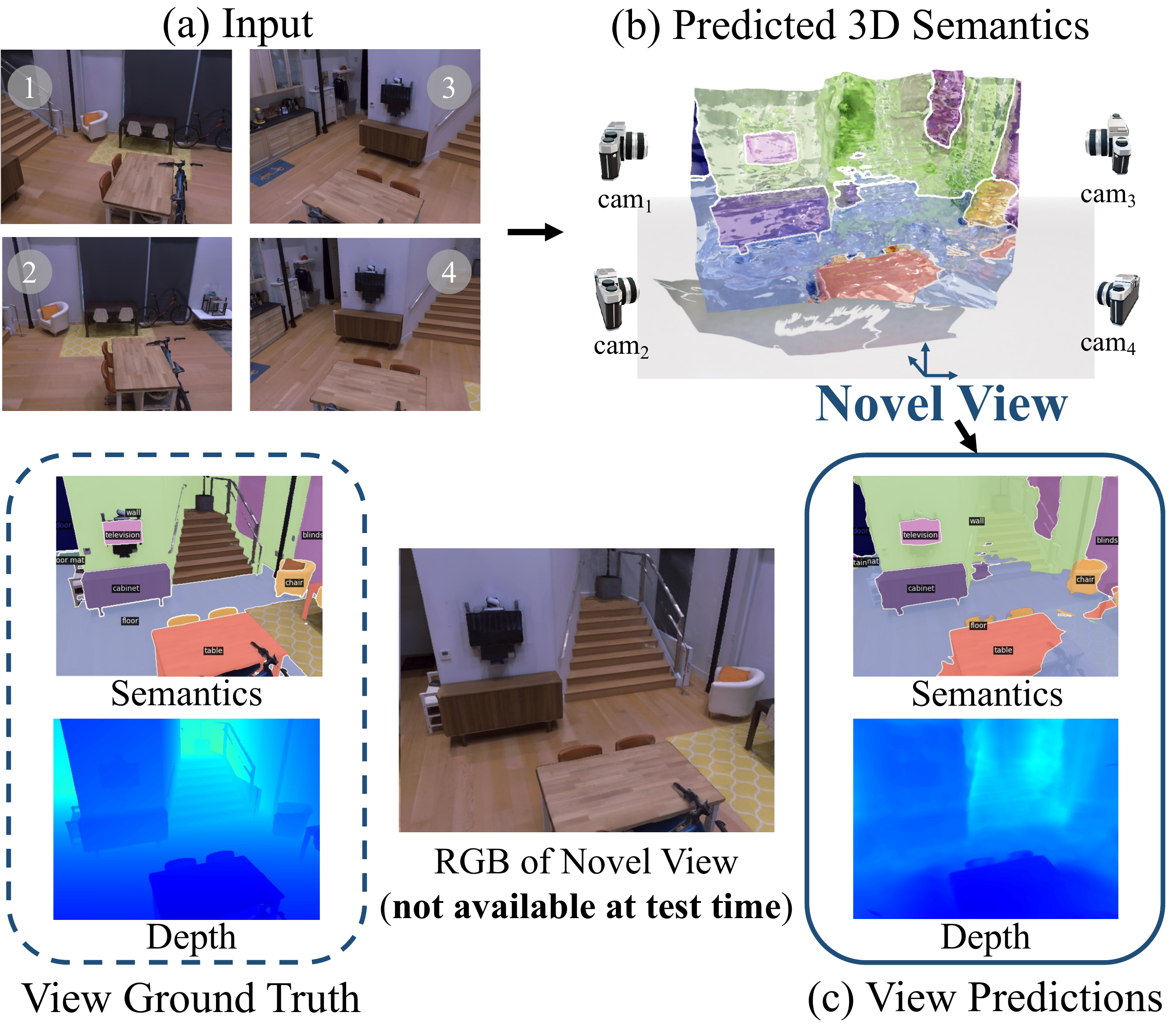}
   \vspace{-6mm}
   \caption{We propose \method which takes as input (a) a few images of a novel scene, and recognizes the scene from novel viewpoints. The novel viewpoint, in the form of camera coordinates, queries (b) our learnt 3D representation to produce (c) semantic segmentations from the view {\it without access to the view's RGB image}. The view query additionally produces (c) depth. \method trains on hundreds of scenes using multi-view 2D annotations and no 3D supervision. Depth colormap: 0.1m \includegraphics[width=0.13\linewidth]{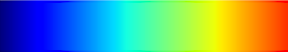} 20m.}
   \label{fig:teaser}
   \vspace{-5mm}
\end{figure}

Humans can build a rich understanding of scenes from a handful of images. The pictures in Fig.~\ref{fig:teaser}, for instance, let us imagine how the objects would occlude, disocclude, and change shape as we walk around the room. This skill is useful in new environments, for example if one were at a friend's house, looking for a table to put a cup on. One can reason that a side table may be by the couch without first mapping the room or worrying about what color the side table is. As one walks into a room, one can readily sense floors behind objects and chair seats behind tables, etc. This ability is so intuitive that entire industries like hotels and real estate depend on persuading users with a few photos. 

The goal of this paper is to give computers the same ability.
In AI, this skill allows autonomous agents to purposefully and efficiently interact with the scene and its objects bypassing the expensive step of mapping.
AR/VR and graphics also benefit from 3D scene understanding. 
To this end, we propose to learn a 3D representation that enables machines to recognize a scene by segmenting it into semantic categories from {\it novel views}.
Each novel view, provided in the form of camera coordinates, queries the learnt 3D representation to produce a semantic segmentation of the scene from that view {\it without access to the view's RGB image}, as we show in Fig.~\ref{fig:teaser}.
We aim to learn this 3D representation without onerous 3D supervision.

Making progress on the problem in complex indoor scenes requires connecting three key areas of computer vision: semantic understanding (i.e., naming objects), 3D understanding, and novel view synthesis (NVS). This puts the problem out of reach of work in any one area. For instance, there have been substantial advances in NVS, including methods like NeRF~\cite{mildenhall2020nerf} or SynSin~\cite{wiles2020synsin}. These approaches perform well on NVS, including results in new scenes~\cite{yu2021pixelnerf,wiles2020synsin}, but are focused on only appearance. In addition to numerous semantic-specific details, recognition in novel viewpoints via direct appearance synthesis is suboptimal: one may be sure of the presence of a rug behind a couch, but unsure of its particular color. Similarly, there have been advances in learning to infer 3D properties of scenes from image cues~\cite{factored3dTulsiani18,Nie_2020_CVPR,Gkioxari2019}, or with differentiable rendering~\cite{kato2018neural,liu2019soft,chen2019learning,ravi2020pytorch3d} and other methods for bypassing the need for direct 3D supervision~\cite{3DRCNN_CVPR18,kanazawa2018learning,kulkarni2020acsm,ye2021shelf}. 
However, these approaches do not connect to complex scene semantics; they primarily focus on single objects or small, less diverse 3D annotated datasets.
More importantly, we empirically show that merely using learned 3D to propagate scene semantics to the new view is insufficient.

We tackle the problem of predicting scene semantics from novel viewpoints, which we refer to as \emph{novel view semantic understanding}, and propose a new end-to-end model, \method.
\method fuses advances and insights from semantic segmentation, 3D understanding, and novel view synthesis with a number of task-specific modifications.
As input, our model takes a few posed RGB images from a previously unseen scene and a target pose (but {\it not} image). As output, it recognizes and segments objects from the target view by producing a semantic segmentation map. As an additional product, it also produces depth in the target view. During training, \method depends only on posed 2D images and semantic segmentations, and in particular does not use ground truth 3D annotations beyond camera pose. 

Our experiments validate \method's contributions on two challenging datasets, Hypersim~\cite{roberts2020hypersim} and Replica~\cite{straub2019replica}. We substantially outperform alternate framings of the problem that build on the state-of-the-art: image-based NVS~\cite{yu2021pixelnerf} followed by semantic segmentation~\cite{chen2018encoder}, which tests whether NVS is sufficient for the task; and lifting semantic segmentations~\cite{chen2018encoder} to 3D and differentiably rendering, like~\cite{wiles2020synsin}, which tests the value of using implicit functions to tackle the problem. 
Our ablations further quantify the value of our problem-specific design. 
Among others, they reveal that \method trained for novel view semantic segmentation obtains more accurate depth predictions compared to a variant which trains without a semantic loss and image-based NVS~\cite{yu2021pixelnerf}, indicating that semantic understanding from novel viewpoints positively impacts geometry understanding.
Overall, our results demonstrate \method's ability to jointly capture the semantics and geometry of unseen scenes when tested on new, complex scenes with diverse layouts, object types, and shapes.

\vspace{-1mm}
\section{Related Work}
\label{sec:related}

For the task of novel view semantic understanding,
we draw from 2D scene understanding, novel view synthesis and 3D learning to recognize scenes from novel viewpoints.

\mypar{Semantic Segmentation.}
Segmenting objects and stuff from images is extensively researched.
Initial efforts apply Bayesian classifiers on local features~\cite{konishi2000statistical} or perform grouping on low-level cues~\cite{shi2000normalized}.
Others~\cite{carreira2012semantic,dai2015convolutional} score bottom-up mask proposals~\cite{arbelaez2014multiscale,carreira2011cpmc}.
With the advent of deep learning,
FCNs~\cite{long2015fully} perform per-pixel segmentation with a CNN.
DeepLab~\cite{chen2017rethinking} use atrous convolutions and an encoder-decoder architecture~\cite{chen2018encoder} to handle scale and resolution.

Regarding multi-view semantic segmentation,
\cite{kundu2016feature} improve the temporal consistency of semantic segmentation in videos by linking frames with optical flow and learned feature similarity.
\cite{mccormac2017semanticfusion} map semantic segmentations from RGBD inputs on 3D reconstructions from SLAM.
\cite{he2017std2p} fuse predictions from video frames using super-pixels and optical flow.
\cite{luc2017predicting} learn scene dynamics to predict semantic segmentations of future
frames given several past frames.

\mypar{Novel View Synthesis.}
Novel view synthesis is a popular topic of research in computer vision and graphics.
\cite{flynn2019deepview,srinivasan2019pushing,srinivasan2017learning,tulsiani2018layer,xu2019deep,zhou2018stereo} show great results synthesizing views from two or more narrow baseline images.
Implicit voxel representations have been used to fit a scene from many scene views~\cite{lombardi2019neural,shin20193d,sitzmann2019deepvoxels}.
Recently, NeRF~\cite{mildenhall2020nerf} learn a continuous volumetric scene function which emits density and radiance at spatial locations and show impressive results when fitted on a single scene with hundreds of views.
We extend NeRF to emit a distribution over semantic categories at each 3D location.
Semantic-NeRF~\cite{ZhiICCV2021} also predicts semantic classes.
We differ from~\cite{ZhiICCV2021} as we generalize to novel scenes from sparse input views instead of in-place interpolation within a single scene from hundreds of views.
NeRF extensions~\cite{henzler2021unsupervised,reizenstein21co3d}, such as PixelNeRF~\cite{yu2021pixelnerf}, generalize to novel scenes from few input views with the help of learnt CNNs but show results on single-object benchmarks for RGB synthesis.
We differ from~\cite{yu2021pixelnerf} by carefully pairing a geometry-aware model with state-of-the-art scene recognition~\cite{chen2018encoder} and experiment on realistic multi-object scenes.

\mypar{3D Reconstruction from Images.}
Scene reconstruction from multiple views is traditionally tackled with classical binocular stereo~\cite{Hartley04, scharstein2002taxonomy} or with the help of shape priors~\cite{Bao2013,blanz1999,dame2013,hane2014}.
Modern techniques learn disparity from image pairs~\cite{kendall2017end}, estimate correspondences with contrastive learning~\cite{schmidt2017self}, perform multi-view stereopsis via differentiable ray projection~\cite{kar2017learning} or learn to reconstruct scenes while optimizing for cameras~\cite{qian2020associative3d,jin2021planar}.
Differentiable rendering~\cite{loper2014opendr,kato2018neural,liu2019soft,chen2019learning,li2018differentiable,nimier2019mitsuba,ravi2020pytorch3d} allows gradients to flow to 3D scenes via 2D re-projections.
\cite{kato2018neural,liu2019soft,chen2019learning,ravi2020pytorch3d} reconstruct single objects from a single view via rendering from 2 or more views during training. 
We also use differentiable rendering to learn 3D via 2D re-projections in semantic space.


\mypar{Depth Estimation from Images.}
Recent methods train networks to predict depth from videos~\cite{zhou2017unsupervised,luo2020consistent,chen2019qanet} or 3D supervision~\cite{eigen2015predicting,chen2016single,Li18,yin2021learning,ranftl2020towards}.
We do not use depth supervision but predict depth from novel views via training for semantic and RGB reconstruction from sparse inputs.

\begin{figure*}[t!]
  \centering
   \includegraphics[width=1.01\linewidth]{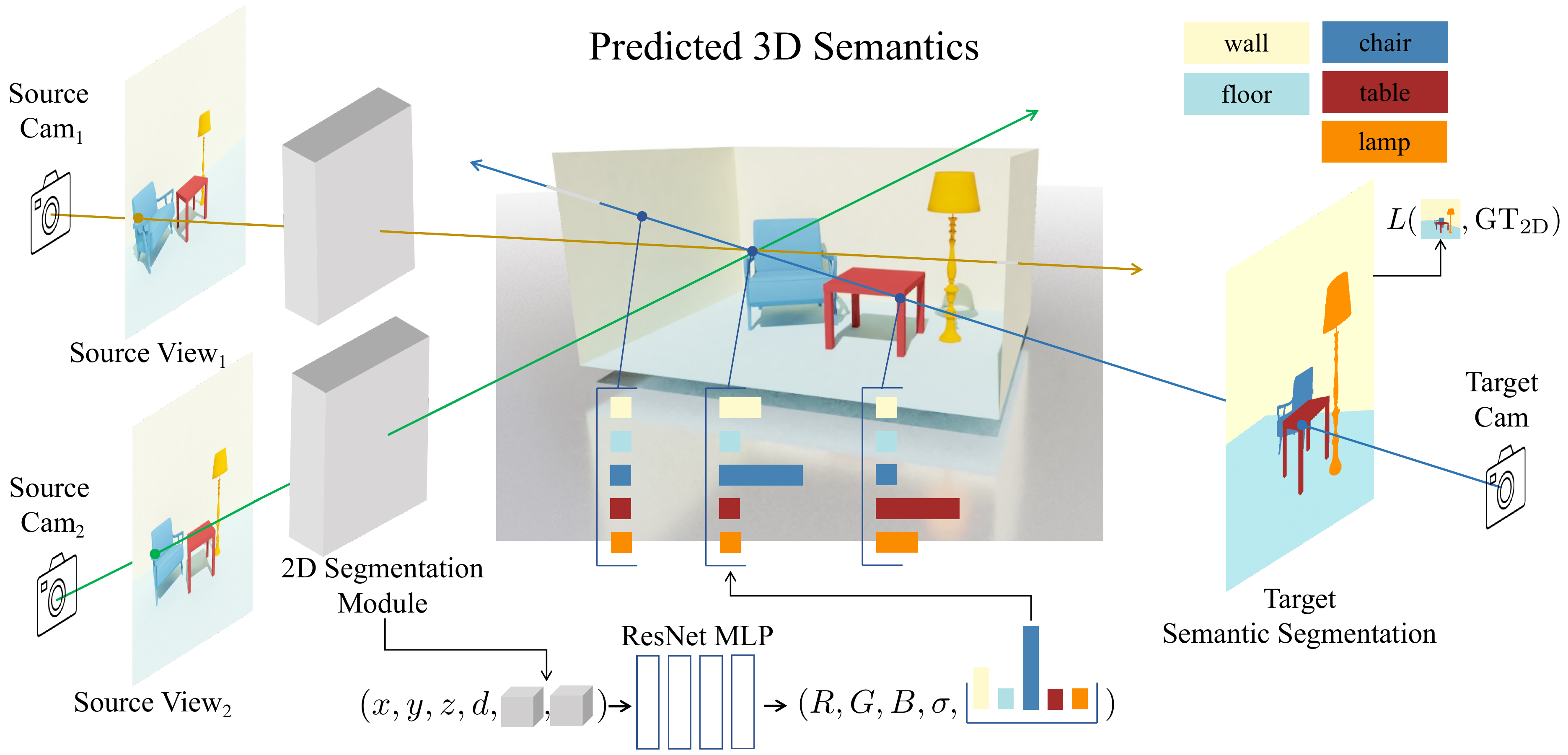}
   \vspace{-8mm}
   \caption{Our model, \method, uses a few source RGB views of a novel scene (here 2) and predicts the semantic segmentation of that scene from a given unseen target viewpoint. 
   Our approach embeds each source view into a latent semantic space with a 2D segmentation module; this space is used to predict radiance, density and distribution over semantic classes at each spatial 3D location via an MLP. 
   The final semantic segmentation is created by volumetric rendering from the target viewpoint. 
   We train with posed multi-view 2D object annotations and no 3D supervision. 
   We generalize to unseen scenes by training on hundreds of diverse scenes and thousands of source-target pairs.}
   \label{fig:overview}
   \vspace{-2mm}
\end{figure*}

\section{Approach}
\label{sec:approach}


We tackle novel view semantic understanding end-to-end with a new model, \method.
\method takes as input RGB images from N \emph{source} views and segments objects and stuff from a novel \emph{target} viewpoint without access to the target image.
We pair semantic segmentation with an implicit 3D representation to learn semantics and geometry from hundreds of scenes and thousands of source-target pairs.
An overview of our approach is shown in Fig.~\ref{fig:overview}.

Our model consists of a 2D semantic segmentation module which embeds each source view to a higher dimensional semantic space. 
An implicit 3D representation~\cite{mildenhall2020nerf} samples features from the output of the segmentation module and predicts radiance, density and a distribution over semantic categories at each spatial 3D location with an MLP.
The 3D predictions are projected to the target view to produce the segmentation from that view via volumetric rendering.

\subsection{Semantic Segmentation Module}

The role of the semantic segmentation module is to project each source RGB view to a learnt feature space, which will subsequently be used to make 3D semantic predictions.
Our 2D segmentation backbone takes as input an image $I$ and outputs a spatial map $M$ of the same spatial resolution, $b^\textrm{seg}: I^{H\times W \times3} \rightarrow M^{H \times W \times K}$, using a convolutional neural network (CNN). Here, $K$ is the dimension of the feature space ($K=256$).

We build on the state-of-the-art for single-view semantic segmentation and follow the encoder-decoder DeepLabv3+~\cite{chen2018encoder,chen2017rethinking} architecture which processes an image 
by downsampling and then upsampling it to its original resolution with a sequence of convolutions.
We remove the final layer, which predicts class probabilities, and use the output from the penultimate layer. 

We initialize our segmentation module by pre-training on ADE20k~\cite{zhou2019semantic}, a dataset of over 20k images with semantic annotations. 
We empirically show the impact of the network architecture and pre-training in our experiments  (Sec.~\ref{sec:experiments}).

\subsection{Semantic 3D Representation}

To recognize a scene from novel viewpoints, we learn a 3D representation which predicts the semantic class for each 3D location.
To achieve this we learn a function $f$ which maps semantic features from our segmentation module to distributions over semantic categories conditioned on the 3D coordinates of each spatial location.

Assume N source views $\{I_j\}_{j=1}^{N}$ and corresponding cameras $\{\pi_j\}_{j=1}^{N}$.
For each view we extract the semantic maps $\{M_j\}_{j=1}^{N}$ with our 2D segmentation module, or $M_j = b^\textrm{seg}(I_j)$.
We project every 3D point $\xB$ to the $j$-th view with the corresponding camera transformation, $\pi_j(\mathbf{x})$, and then sample the $K$-dimensional feature vector from $M_j$ from the projected 2D location. 
This yields a semantic feature from the $j$-th image denoted as $\phi^\textrm{seg}_j(\mathbf{x}) = M_j(\pi_j(\mathbf{x}))$.

Following NeRF~\cite{mildenhall2020nerf} and PixelNeRF~\cite{yu2021pixelnerf}, $f$ takes as input a positional embedding of the 3D coordinates of $\mathbf{x}$, $\gamma(\mathbf{x})$, and the viewing direction $\mathbf{d}$. 
We additionally feed the semantic embeddings $\{\phi^\textrm{seg}_j\}_{j=1}^{N}$. 
As output, $f$ produces
\begin{equation}
    (\cB, \, \sigma, \, \sB) = f\bigl( \gamma(\mathbf{x}), \, \mathbf{d}, \, \phi^\textrm{seg}_0(\mathbf{x}), \, ..., \, \phi^\textrm{seg}_{N-1}(\mathbf{x}) \bigr)
\end{equation}
where $\cB \in \mathbb{R}^3$ is the RGB color, $\sigma \in \mathbb{R}$ is the occupancy, and $\sB \in \mathbb{R}^{|\mathcal{C}|}$ is the distribution over semantic classes $\mathcal{C}$.

We model $f$ as a fully-connected ResNet~\cite{he2016deep}, similar to PixelNeRF~\cite{yu2021pixelnerf}.
The positional embedding of the 3D location and the direction are concatenated to each semantic embedding $\phi^\textrm{seg}_j$, and each is passed through 3 residual blocks with 512 hidden units.
The outputs are subsequently aggregated using average pooling and used to predict the final outputs of $f$ via two branches: one predicts the semantics $\sB$, the other the color $\cB$ and density $\sigma$. 
Each branch consists of two residual blocks, each with 512 hidden units.
Read about the network architecture in the Supplementary.

\mypar{Predicting Semantics.}
Rendering the semantic predictions, $s$, from a given viewpoint gives the semantic segmentation of the scene from that view.
Following NeRF~\cite{mildenhall2020nerf}, we accumulate predictions on rays, $\mathbf{r}(t) = \mathbf{o} + t \cdot \mathbf{d}$, originating at the camera center $\mathbf{o}$ with direction $\mathbf{d}$, 
\begin{equation}
    \hat{S}(\mathbf{r}) = \int_{t_n}^{t_f} T(t) \sigma(t) \sB(t) dt
    \label{eq:semantic}
\end{equation}
where $T(t) = \exp(- \int_{t_n}^{t} \sigma(s) ds)$ is the accumulated transmittance along the ray, and $t_n$ and $t_f$ are near and far sampling bounds, which are hyperparameters. 

The values of the sampling bounds $(t_n, t_f)$ are crucial for good performance. 
In the original NeRF~\cite{mildenhall2020nerf} method, $(t_n, t_f)$ are manually set to tightly bound the scene.
PixelNeRF~\cite{ZhiICCV2021} uses manually selected parameters \emph{for each object category}, \ie different values for chairs, different values for cars, \etc.
In Semantic-NeRF~\cite{ZhiICCV2021}, the values are selected for the Replica rooms, which vary little in size.
In the datasets we experiment on in Sec.~\ref{sec:experiments}, scene scale varies drastically from human living spaces with regular depth extents (\eg living rooms) to industrial facilities (\eg warehouses), lofts or churches with large far fields.
With the goal of scene generalization in mind, we set $(t_n, t_f)$ globally, regardless of the true near/far bounds of each scene we encounter.
This more realistic setting makes the problem harder: our model needs to predict the right density values for a large range of depth fields, reasoning about occupancy within each scene but also about the depth extent of the scene as a whole.

Replacing the semantic predictions $\sB$ with the RGB predictions $\cB$ in Eq.~\ref{eq:semantic} produces the RGB view $\hat{C}(\mathbf{r})$ from the target viewpoint, as in~\cite{mildenhall2020nerf}.
While photometric reconstruction is not our goal, we use $\hat{C}$ during learning and show that it helps capture the scene's geometry more accurately. 

\mypar{Predicting Depth.}
In addition to the semantic segmentation $\hat{S}$ and the RGB reconstruction $\hat{C}$ from a target viewpoint, we also predict the pixel-wise depth of the scene from that view, as in~\cite{deng2021depth} by computing
\begin{equation}
    \hat{D}(\mathbf{r}) = \int_{t_n}^{t_f} T(t) \sigma(t) t dt.
    \label{eq:depth}
\end{equation}
We use the depth output $\hat{D}$ only during evaluation. 
By comparing single-view depth ($\hat{D}$) and semantic segmentation ($\hat{S}$) from many novel viewpoints, we measure our model's ability to capture geometry and semantics.

\subsection{Learning Objective}
\label{sec:optimization}

Our primary objective is to segment the scene from the target view. We jointly train the segmentation module $b^\textrm{seg}$ and implicit 3D function $f$ to directly solve this task. We also find that auxiliary photometric and source-view losses are crucial for performance. Our objectives require RGB images and 2D semantics from various views in the scene as well as poses to perform re-projection. 



Since our goal is to predict a semantic segmentation in the target view, our primary objective is a per-pixel cross-entropy loss between the true class labels $S$ and predicted class distribution $\hat{S}$, 
\begin{equation}
    L^S_{\textrm{target}} = - \sum_{\mathbf{r} \in \mathbf{R}} \sum_{j=1}^{|\mathcal{C}|} S^j(\mathbf{r}) \log \hat{S}^j (\mathbf{r})
\end{equation}
where $\mathcal{C}$ is the set of semantic classes and $\mathbf{R}$ is the set of rays in the target view. Here, $S^j(\rB)$ is the $\{0,1\}$ true label for the $j$-th class at the intersection of ray $\rB$ and the image.

In addition to this, we minimize auxiliary losses that improve performance on our primary task. The first is a photometric loss on RGB images, namely the squared L$_2$ distance between the prediction $\hat{C}$ and actual image $C$, or 
\begin{equation}
    L^P_{\textrm{target}} = \sum_{\mathbf{r} \in \mathbf{R}}  \left|\left| \hat{C}(\mathbf{r}) - C(\mathbf{r}) \right|\right|^2_2 
\end{equation}
where $C(\mathbf{r})$ is the true RGB color at the intersection of ray $\rB$ and the image.

Finally, in addition to standard losses on the target view~\cite{mildenhall2020nerf,yu2021pixelnerf,ZhiICCV2021}, we find it is important to apply losses on the source views.
Specifically, we create $L_{\textrm{source}}^S$ and $L_{\textrm{source}}^P$ that are the semantic and photometric losses, respectively, applied to rays from the source views. These losses help enforce consistency with the input views.
Our final objective is given by
\begin{equation}
    L = L^P_{\textrm{target}} + L^P_{\textrm{source}} + \lambda \cdot (L^S_{\textrm{target}}  + L^S_{\textrm{source}})
    \label{eq:objective}
\end{equation}
where $\lambda$ scales the semantic and photometric losses.

\section{Experiments}
\label{sec:experiments}

We experiment on Hypersim~\cite{roberts2020hypersim} and Replica~\cite{straub2019replica}.
Both provide posed views of complex scenes with over 30 object types and under varying conditions of occlusion and lighting.
At test time, we evaluate on novel scenes not seen during training.
Due to its large size, we treat Hypersim as our main dataset where we run an extensive quantitative analysis.
We then show generalization to the smaller Replica.

\mypar{Metrics.}
We report novel view metrics for semantics and geometry. 
For a novel view of a test scene, we project the semantic predictions (Eq.~\ref{eq:semantic}) and depth (Eq.~\ref{eq:depth}) and compare to the ground truth semantic and depth maps, respectively.
Ideally, we would also evaluate directly in 3D, which requires access to full 3D ground truth.
However, 3D ground truth is not publicly available for Hypersim and is generally hard to collect.
Thus, we treat novel view metrics as proxy metrics for 3D semantic segmentation and depth estimation.

For semantic comparisons, we report semantic segmentation metrics~\cite{caesar2018cvpr} implemented in Detectron2~\cite{wu2019detectron2}:
\textbf{mIoU} is the intersection-over-union (IoU) averaged across classes, 
\textbf{IoU$^\textrm{T}$} and \textbf{IoU$^\textrm{S}$} report IoU by merging all things (object) and stuff classes (wall, floor, ceiling), respetively.
\textbf{fwIoU} is the per class IoU weighted by the pixel-level frequency of each classs, 
\textbf{pACC} is the percentage of correctly labelled pixels and 
\textbf{mACC} is the pixel accuracy averaged across classes. 
For all, performance is in \% and higher is better.

For depth comparisons, we report depth metrics following~\cite{Eigen14}:
\textbf{L$_1$} is the per-pixel average L$_1$ distance between ground truth and predicted depth, 
\textbf{Rel} is the L$_1$ distance normalized by the true depth value, 
\textbf{Rel$^\textrm{T}$} and \textbf{Rel$^\textrm{S}$} is the Rel metric for all things and stuff, respectively.
$\bm{\delta < \tau}$ is the percentage of pixels with predicted depth within $[\frac{1}{\tau}, \tau] \times$ the true depth. 
L$_1$ is in meters and $\delta < \tau$ is in \%.
For $\delta < \tau$ metrics, higher is better. 
For all other, lower is better ($\downarrow$).

\subsection{Experiments on Hypersim}

Hypersim~\cite{roberts2020hypersim} is a dataset of 461 complex scenes.
Camera trajectories across scenes result in 77,400 images with camera poses, masks for 40 semantic classes~\cite{Silberman12}, along with true depth maps. 
Hypersim contains on average 50 objects per image, making it a very challenging dataset.

\mypar{Dataset.}
For each scene, we create source-target pairs from the available views. 
Each image is labelled as target and is paired with an image from a different viewpoint if: (1) the view frustums intersect by no less than 10\%; (2) the camera translation is greater than 0.5m; and (3) the camera rotation is at least 30$^o$.
This ensures that source and target views are from different camera viewpoints and broadly depict the same parts of the scene but without large overlap. 
We follow the original Hypersim split, which splits train/val/test to 365/46/50 disjoint scenes, respectively.
Overall, there are 120k/14k/14k pairs in train/val/test, respectively.

\begin{figure*}[t]
  \centering
   \includegraphics[width=0.99\linewidth]{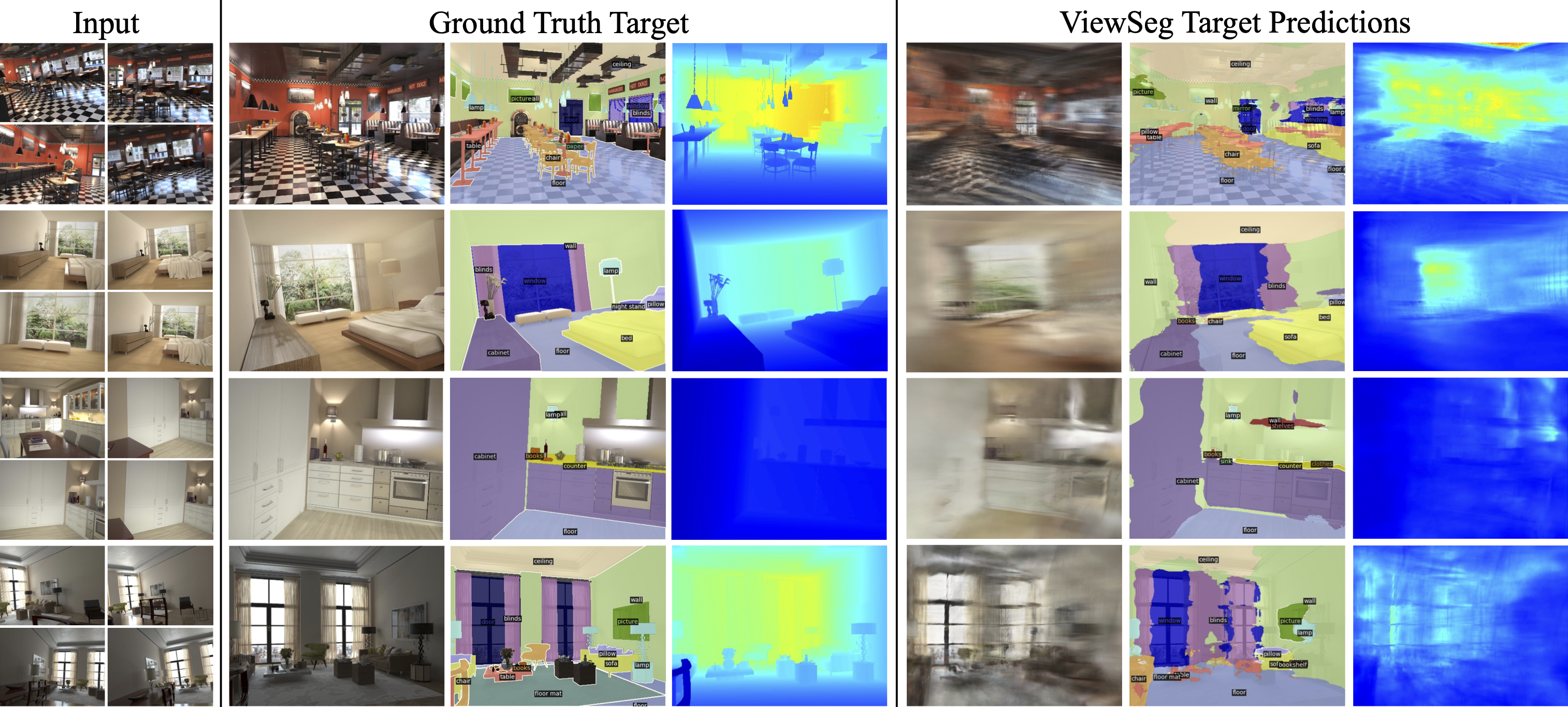}
   \vspace{-3mm}
   \caption{Predictions on Hypersim.
   For each example, we show the 4 input RGB views (left), the ground truth RGB, semantic and depth maps for the novel target view (middle) and \method's predictions (right).
   RGB synthesis is not our goal, but we show the predicted RGB for completeness. Our model does not have access to the true observations from the target view at test time. Depth: 0.1m \includegraphics[width=0.07\linewidth]{figures/depth_colormap.png} 20m.}
   \label{fig:hypersim_preds}
   \vspace{-2mm}
\end{figure*}
\begin{table*}[h!]
   \centering
   \scalebox{0.85}{
   \begin{tabular}{L{.13\linewidth}|S{.05\linewidth}S{.05\linewidth}S{.05\linewidth}S{.05\linewidth}S{.05\linewidth}S{.05\linewidth}|D{.05\linewidth}D{.07\linewidth}D{.07\linewidth}D{.07\linewidth}D{.08\linewidth}D{.09\linewidth}D{.09\linewidth}}
	Model & mIoU & IoU$^\textrm{T}$ & IoU$^\textrm{S}$ & fwIoU  & pACC & mACC  & L$_1$ ($\downarrow$) & Rel ($\downarrow$) &Rel$^\textrm{T}$ ($\downarrow$) & Rel$^\textrm{S}$ ($\downarrow$) & $\delta < 1.25$ & $\delta < 1.25^2$ \\
	\hhline
	PixelNeRF++ & 1.58 & 14.9 & 47.9 & 17.9 & 3.63 & 35.9 & 2.80 & 0.746 & 0.856 & 0.653 & 0.300 & 0.531 \\ 
	\cloud & 0.46 & 29.6 & 4.42 & 1.80 & 3.31 & 3.25 & 3.81 & 0.856 & 0.997 & 0.737 & 0.145 & 0.277  \\
	\method & \bf{17.1} & \bf{33.2} & \bf{58.9} &  \bf{44.8} & \bf{23.9} & \bf{62.2} & \bf{2.29} & \bf{0.646} & \bf{0.721} & \bf{0.584} & \bf{0.409} & \bf{0.656} \\
	\hline
	Oracle & 40.0 & 58.1 & 71.3 & 66.6 & 52.1 & 79.1 & 0.96 & 0.235 & 0.317 & 0.163 & 0.731 & 0.898 \\ 
  \end{tabular}
  } 
  \vspace{-3mm}
  \caption{Comparisons on Hypersim val. We report the performance of both semantic segmentation (\textcolor{llblue}{blue}) and depth estimation (\textcolor{llgreen}{green}) for our method, \method, an oracle which applies supervised single-image semantic segmantation and depth estimation models on the true target RGB views, and two competing approaches, PixelNeRF++ and an explicit 3D point cloud model, \cloud, inspired by SynSin~\cite{wiles2020synsin}.}
  \label{tab:hypersim_comp}
  \vspace{-3mm}
\end{table*}

\mypar{Training details.}
We implement \method in PyTorch with Detectron2~\cite{wu2019detectron2} and PyTorch3D~\cite{ravi2020pytorch3d}.
We train on the Hypersim training set for 13 epochs with a batch size of 32 across 32 Tesla V100 GPUs.
The input and render resolution are set to 1024$\times$768, maintaining the size of the original dataset.
We optimize with Adam~\cite{kingma2014adam} and a learning rate of 5e-4.
We follow the PixelNeRF~\cite{yu2021pixelnerf} strategy for ray sampling: We sample 64 points per ray in the coarse pass and 128 points per ray in the fine pass.
In addition to the target view, we randomly sample rays on each source view and additionally minimize the source view loss, as we describe in Sec.~\ref{sec:optimization}.
We set $t_n = 0.1$m, $t_f=20$m in Eq.~\ref{eq:semantic} \&~\ref{eq:depth} and $\lambda = 0.04$ in Eq.~\ref{eq:objective}. 
More details in the Supplementary.

\mypar{Baselines.}
In addition to extensive ablations that reveal which components of our method are most important, we compare with multiple baselines and oracle methods to provide context for our method. 
Our baselines aim to test alternate strategies, including inferring the true RGB image and then predicting the pixel classes as well as lifting a predicted semantic segmentation map to 3D and re-projecting.

To provide context, we report a {\bf Target View Oracle} that has access to the true target view's image. 
The target RGB image fundamentally resolves many ambiguities in 3D about what is where, and is not available to our method. 
Instead, our method is tasked with predicting segmentations and depth from new viewpoints \emph{without the target RGB images}.
Our oracle applies appropriate supervised models directly on the true target RGB. 
For semantic segmentation, we use a model~\cite{chen2018encoder} that is identical to ours pre-trained on ADE20k~\cite{zhou2019semantic} and finetuned on all images of the Hypersim training set. 
For depth, we use the model from~\cite{yin2021learning}, which predicts normalized depth.
We obtain metric depth by aligning with the optimal shift and scale following \cite{ranftl2020towards,yin2021learning}.

Our first baseline, denoted {\bf PixelNeRF++}, tests the importance of the end-to-end nature of our approach by performing a two stage process: we use the novel view synthesis (NVS) method of PixelNeRF~\cite{yu2021pixelnerf} to infer the RGB of the target view and then apply an image-based model for semantic segmentation. 
To ensure a fair comparison, PixelNeRF is trained on the Hypersim training set.
We use the segmentation model we trained for the oracle to predict semantic segmentations. 
Depth is predicted with Eq.~\ref{eq:depth}.


Our second baseline, named {\bf \cloud}, tests the importance of an implicit representation by comparing with an explicit 3D point cloud representation. 
Inspired by SynSin~\cite{wiles2020synsin}, we train a semantic segmentation backbone similar to our \method, along with a depth model, from~\cite{yin2021learning}, to lift each source view to a 3D point cloud with per point class probabilities. 
A differentiable point cloud renderer~\cite{ravi2020pytorch3d} projects the point clouds from the source images to the target view to produce a semantic and a depth map.
\cloud is trained on the Hypersim training set and uses the same 2D supervision as our \method.

\begin{figure}[t]
  \centering
   \includegraphics[width=1.0\linewidth]{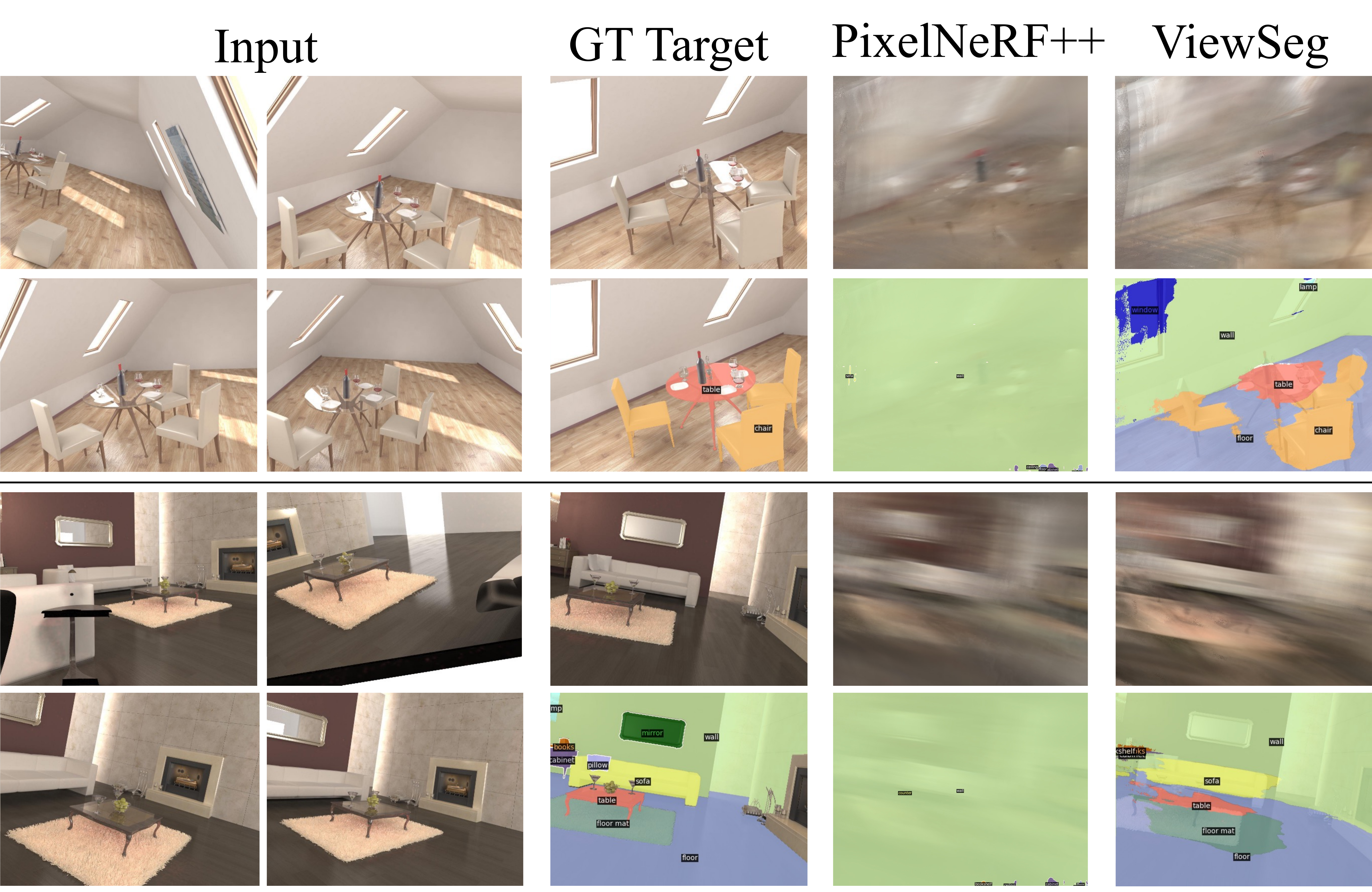}
   \vspace{-6mm}
   \caption{Comparison of \method and PixelNeRF++~. For each example we show the 4 RGB inputs (1$^\textrm{st}$-2$^\textrm{nd}$ col.), the true RGB and semantic map from the target view (3$^\textrm{rd}$ col.), the RGB and semantic prediction by PixelNeRF++ (4$^\textrm{th}$ col.) and the RGB and semantic prediction by our \method (5$^\textrm{th}$ col.).}
   \label{fig:hypersim_comp}
\end{figure}

\mypar{Results.}
Table~\ref{tab:hypersim_comp} compares our \method, PixelNeRF++ and \cloud with 4 source views and the oracle on Hypersim val.
We observe that PixelNeRF++, which predicts the target RGB view and then applies an image-based model for semantic prediction, performs worse than \method. 
This is explained by the low quality RGB predictions, shown in Fig.~\ref{fig:hypersim_comp}.
Predicting high fidelity RGB of novel complex scenes from only 4 input views is still difficult for NVS models, suggesting that a two-stage solution for semantic segmentation will not perform competitively.
Indeed, \method significantly outperforms PixelNeRF++ showing the importance of learning semantics end-to-end. 
In addition to semantics, \method outperforms PixelNeRF++ for depth.
This suggests that learning semantics jointly has a positive impact on geometry as well.
Finally, \cloud has a hard time predicting semantics and geometry.
This is likely attributed to the wide baseline source and target views in our task which cause explicit 3D representations to produce holes and erroneous predictions in the rendered target output.
In SynSin~\cite{wiles2020synsin}, the camera transform between source and target in the datasets the authors explore is significantly narrower, unlike our task where novel viewpoints correspond to wider camera transforms, as shown in Fig.~\ref{fig:hypersim_preds}.

Fig.~\ref{fig:hypersim_preds} shows \method's predictions on Hypersim val. 
We show the 4 source views (left), the ground truth target RGB, semantic and depth map (middle) and \method's predictions from the target viewpoint (right).
Note that \method does not have access to ground truth target observations and only receives the 4 images along with camera coordinates for the source and the target viewpoints.
Examples in Fig.~\ref{fig:hypersim_preds} are of diverse scenes (restaurant, bedroom, kitchen, living room) with many objects (chair, table, counter, cabinet, window, blinds, lamp, picture, floor mat, \etc).
We observe that the predicted RGB is of poor quality proving that NVS has a hard time in complex scenes and with few views. 
RGB synthesis is not our goal.
We aim to predict the scene semantics and Fig.~\ref{fig:hypersim_preds} shows that our model achieves this. 
\method detects stuff (floor, wall, ceiling) well and predicts object segments for the right object types, even for diverse target views.
Our depth predictions show that \method captures the scene's geometry, even though it was not trained with any 3D supervision.
 
\begin{figure}[h!]
  \centering
   \includegraphics[width=1.0\linewidth]{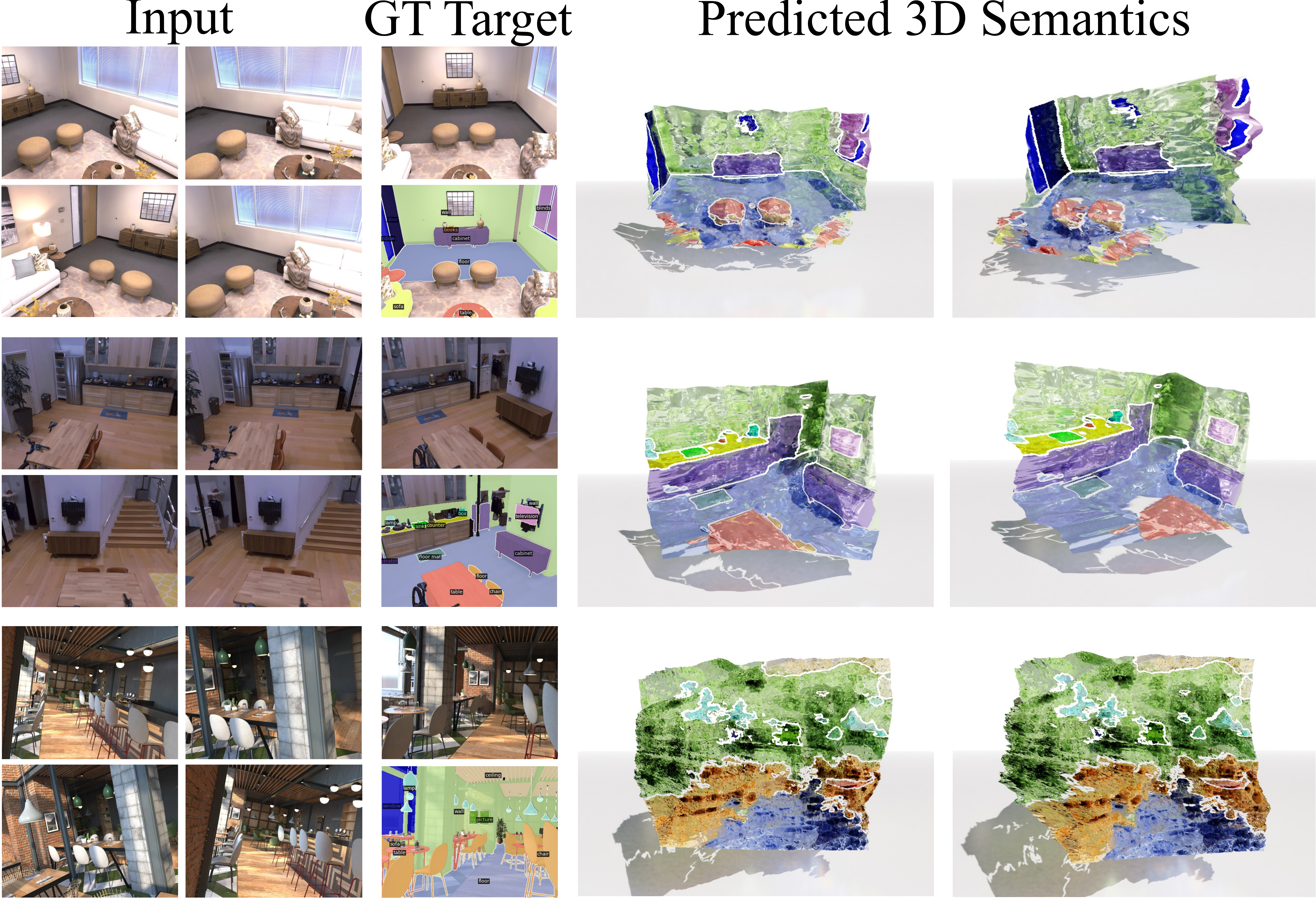}
   \vspace{-5mm}
   \caption{3D semantic reconstructions on Replica (top two) and Hypersim (bottom). We show the 4 RGB inputs (1$^\textrm{st}$-2$^\textrm{nd}$ col.), the true RGB and semantic map from the novel view (3$^\textrm{rd}$ col.), and two views of the 3D semantic reconstructions (4$^\textrm{th}$-5$^\textrm{th}$ col.).}
   \label{fig:results_3d}
   \vspace{-4mm}
\end{figure}

Fig.~\ref{fig:hypersim_comp} compares \method to PixelNeRF++. 
Semantic segmentation from predicted RGB results in bad predictions, as shown in the PixelNeRF++ column.
Fig.~\ref{fig:results_3d} shows examples of semantic 3D reconstructions.

\begin{table*}[h!]
   \centering
   \scalebox{0.86}{
   \begin{tabular}{L{.17\linewidth}|S{.04\linewidth}S{.04\linewidth}S{.04\linewidth}S{.05\linewidth}S{.05\linewidth}S{.05\linewidth}|D{.05\linewidth}D{.06\linewidth}D{.07\linewidth}D{.07\linewidth}D{.08\linewidth}D{.09\linewidth}D{.09\linewidth}}
	\method loss & mIoU & IoU$^\textrm{T}$ & IoU$^\textrm{S}$ & fwIoU  & pACC & mACC  & L$_1$ ($\downarrow$) & Rel ($\downarrow$) & Rel$^\textrm{T}$ ($\downarrow$) & Rel$^\textrm{S}$ ($\downarrow$) & $\delta < 1.25$ & $\delta < 1.25^2$ \\
	\hhline
	\emph{w/o} photometric loss  & 16.9 & 30.8 & 58.7 & 44.8 & 22.7 & 62.5 & 2.49 & 0.677 & 0.750 & 0.615 & 0.359 & 0.611 \\
	\emph{w/o} semantic loss     & - & - & - & - & - & - & 2.58 & 0.787 & 0.919 & 0.678 & 0.345 & 0.587 \\
	\emph{w/o} source view loss       & 14.3 & 28.2 & 57.9 & 28.2 & 19.3 & 61.1 & 2.37 & 0.683 & 0.764 & 0.615 & 0.397 & 0.649 \\
    \emph{w/o} viewing dir       & 16.0 & 33.1 & \bf{59.2} & \bf{44.9} & 21.5 & 62.1 & 2.53 & 0.708 & 0.783 & 0.646 & 0.354 & 0.602 \\
    final                        & \bf{17.1} & \bf{33.2} & 58.9 & 44.8 & \bf{23.9} & \bf{62.2} & \bf{2.29} & \bf{0.646} & \bf{0.721} & \bf{0.584} & \bf{0.409} & \bf{0.656}\\
  \end{tabular}
  } 
  \vspace{-3mm}
  \caption{Ablating loss components. We report semantic (\textcolor{llblue}{blue}) and depth (\textcolor{llgreen}{green}) performance on Hypersim val without the photometric, the semantic and the source view loss and when excluding the viewing direction from the input. Our model is reported in the last row.}
  \label{tab:hypersim_loss_ablations}
  \vspace{-2mm}
\end{table*}

\begin{table*}[h!]
   \centering
   \scalebox{0.87}{
   \begin{tabular}{L{.22\linewidth}|S{.05\linewidth}S{.05\linewidth}S{.05\linewidth}S{.05\linewidth}S{.05\linewidth}S{.05\linewidth}|D{.05\linewidth}D{.07\linewidth}D{.07\linewidth}D{.07\linewidth}D{.08\linewidth}D{.09\linewidth}}
	\method backbone & mIoU & IoU$^\textrm{T}$ & IoU$^\textrm{S}$ & fwIoU  & pACC & mACC  & L$_1$ ($\downarrow$) &  Rel ($\downarrow$) & Rel$^\textrm{T}$ ($\downarrow$) & Rel$^\textrm{S}$ ($\downarrow$) & $\delta < 1.25$  \\
	\hhline
	DLv3+~\cite{chen2018encoder} + ADE20k~\cite{zhou2019semantic} & \bf{17.1} & \bf{33.2} & 58.9 & 44.8 & \bf{23.9} & 62.2 & 2.29 & 0.645 & 0.721 & 0.584 & 0.409 \\
    DLv3+~\cite{chen2018encoder} + IN~\cite{deng2009imagenet} & 16.3 & \bf{33.2} & \bf{59.2} & \bf{45.2} & 22.0 & \bf{62.5} & \bf{2.28} & \bf{0.614} & \bf{0.682} & \bf{0.559} & \bf{0.415} \\
    ResNet34~\cite{he2016deep} + IN~\cite{deng2009imagenet} & 7.45 & 21.7 & 55.9 & 37.1 & 11.2 & 56.1 & 2.67 & 0.712 & 0.815 & 0.626 & 0.320 \\
  \end{tabular}
  } 
  \vspace{-2mm}
  \caption{Performance on Hypersim val with different semantic segmentation backbones. We show the performance of \method with DeepLabv3+ (DLv3+)~\cite{chen2018encoder} pretrained on ADE20k~\cite{zhou2019semantic} and on ImageNet~\cite{deng2009imagenet} and a ResNet34~\cite{he2016deep} backbone pretrained on ImageNet~\cite{deng2009imagenet}. 
  The latter is used in PixelNeRF~\cite{yu2021pixelnerf}.
  DLv3+ improves performance significantly, while ADE20k helps ever so slightly.}
  \label{tab:hypersim_backbone_ablations}
  \vspace{-4mm}
\end{table*}

\begin{table}[t]
   \centering
   \scalebox{0.95}{
   \begin{tabular}{L{.2\linewidth}|S{.1\linewidth}S{.1\linewidth}|D{.13\linewidth}D{.13\linewidth}}
	\method & mIoU & pACC & L$_1$ ($\downarrow$) & Rel ($\downarrow$) \\
	\hhline
	\emph{w/} 4 views & \bf{17.1} & \bf{23.9} & \bf{2.29} & \bf{0.584} \\
    \emph{w/} 3 views & 15.5 & 20.8 & 2.39 & 0.652 \\ 
	\emph{w/} 2 views & 13.6 & 18.2 & 2.57 & 0.765 \\ 
	\emph{w/} 1 view  & 11.6 & 15.8 & 2.62 & 0.734 \\
  \end{tabular}
  } 
  \vspace{-2mm}
  \caption{Input study on Hypersim val for varying number of source views. More views improve semantic and depth performance.}
  \label{tab:hypersim_views_ablations}
  \vspace{-5mm}
\end{table}

\mypar{Ablations and Input Study.}
Table~\ref{tab:hypersim_loss_ablations} ablates various terms in our objective.
For reference, the performance of our \method trained with 4 source views and with the final objective (Eq.~\ref{eq:objective}) is shown in the last row.
When we remove the photometric loss, L$^\textrm{P}$, semantic performance remains roughly the same but depth performance drops ($-20$cm in L$_1$), which proves that appearance helps capture scene geometry.
When we remove the semantic loss, L$^\textrm{S}$, and train solely with a photometric loss, we observe a drop in depth ($-29$cm in L$_1$).
This suggests that semantics helps geometry; we made a similar observation when comparing to PixelNeRF++ in Table~\ref{tab:hypersim_comp}.
When training without source view losses both semantic and depth performance drop, with semantic performance deteriorating the most ($-2.8$\% in mIoU). 
This confirms our insight that enforcing consistency with the source views improves learning.
Finally, when we remove the viewing direction from the model's input, depth performance suffers the most ($-24$cm in L$_1$).

Table~\ref{tab:hypersim_backbone_ablations} compares different backbones for the 2D segmentation module.
We compare DeepLabv3+ (DLv3+)~\cite{chen2018encoder} pre-trained on ImageNet~\cite{deng2009imagenet} and ADE20k~\cite{zhou2019semantic} and ResNet34~\cite{he2016deep} pre-trained on ImageNet~\cite{deng2009imagenet}. 
The latter is used in PixelNeRF~\cite{yu2021pixelnerf}.
DLv3+ significantly boosts performance for both semantics and depth while pre-training on ADE20k slightly adds to the final performance.

Table~\ref{tab:hypersim_views_ablations} compares \method with varying number of source views. 
We observe that more views improve both semantic segmentation and depth.
More than 4 views could lead to further improvements but substantially increase memory and time requirements during training.

\begin{figure*}[t!]
  \centering
   \includegraphics[width=0.99\linewidth]{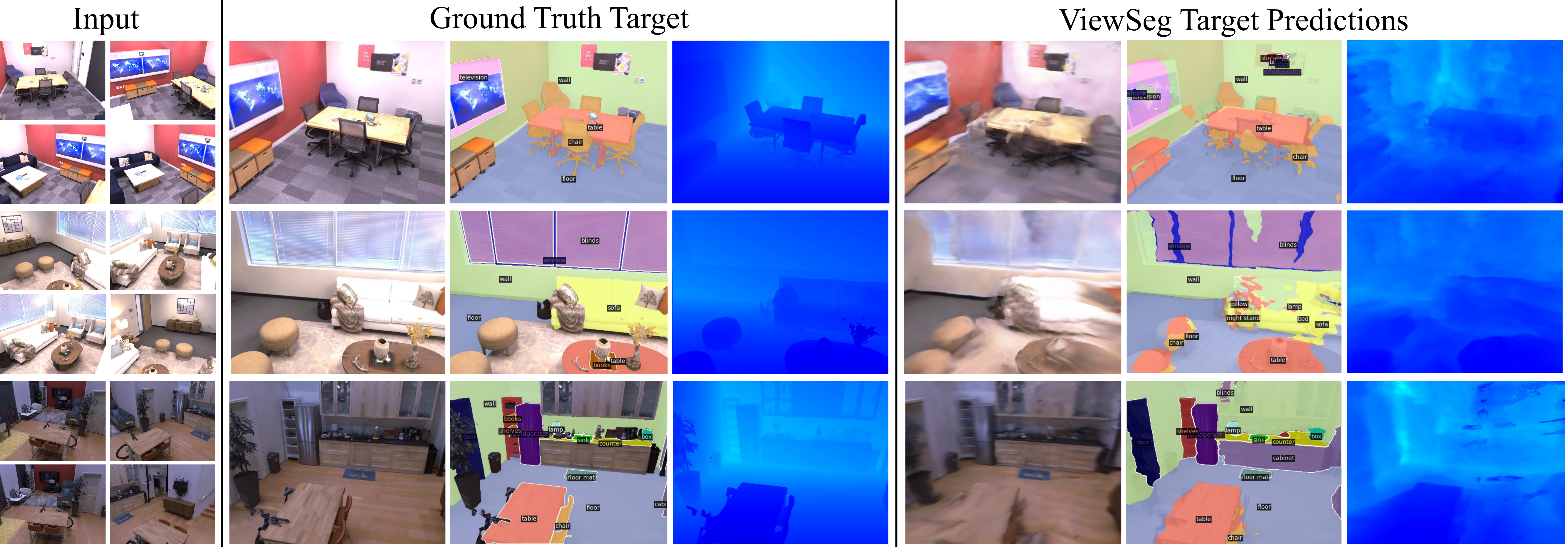}
   \vspace{-2mm}
   \caption{Results on Replica test. We show the 4 input views (left), the ground truth RGB, semantic and depth maps from the novel target viewpoint (middle) and \method's predictions (right). Depth colormap: 0.1m \includegraphics[width=0.07\linewidth]{figures/depth_colormap.png} 20m.}
   \label{fig:results_replica}
   \vspace{-2mm}
\end{figure*}

\begin{table*}
   \centering
   \scalebox{0.86}{
   \begin{tabular}{L{.13\linewidth}|S{.05\linewidth}S{.05\linewidth}S{.05\linewidth}S{.05\linewidth}S{.05\linewidth}S{.05\linewidth}|D{.05\linewidth}D{.07\linewidth}D{.07\linewidth}D{.07\linewidth}D{.08\linewidth}D{.09\linewidth}D{.09\linewidth}}
	Model & mIoU & IoU$^\textrm{T}$ & IoU$^\textrm{S}$ & fwIoU  & pACC & mACC  & L$_1$ ($\downarrow$) & Rel ($\downarrow$) & Rel$^\textrm{T}$ ($\downarrow$) & Rel$^\textrm{S}$ ($\downarrow$) & $\delta < 1.25$ & $\delta < 1.25^2$ \\
	\hhline
	\method noft & 13.2 & 44.8 & 56.0 & 51.4 & 27.1 & 66.8 & 0.982 & 0.222 & 0.194 & 0.254 & 0.623 & 0.880 \\
	\method & 30.2 & 56.2 & 62.8 & 62.3 & 48.4 & 75.6 & 0.550 & 0.130 & 0.130 & 0.130 & 0.851 & 0.961 \\
	\hline
	Oracle & 56.2 & 76.8 & 78.0 & 90.1 & 79.4 & 93.8 & 0.226 & 0.058 & 0.065 & 0.050 & 0.976 & 0.998\\
  \end{tabular}
  } 
  \vspace{-2mm}
  \caption{Performance for semantic segmentation (\textcolor{llblue}{blue}) and depth (\textcolor{llgreen}{green}) on the Replica~\cite{straub2019replica} test set before finetuning (noft) and after finetuning \method on Replica's training set. We additionally report the target view oracle.}
  \label{tab:replica_comp}
\end{table*}

\subsection{Generalization to Replica}

We experiment on the Replica dataset~\cite{straub2019replica} which contains real-world scenes such as living rooms and offices. 
Scenes are complex with many object types and instances in various layouts. 
We show generalization by applying \method pre-trained on Hypersim and then further fine-tune it on Replica to better fit to Replica's statistics.

\mypar{Dataset.}
We use AI-Habitat \cite{savva2019habitat} to extract multiple views per scene. 
For each view we collect the RGB image, semantic labels and depth.
For each Replica scene, we simulate an agent circling around the center of the 3D scene and render the observations.
Note that this is unlike Hypersim~\cite{roberts2020hypersim}, where camera trajectories are extracted by the authors a-priori. 
We use the same camera intrinsics and resolution as Hypersim:
the horizontal field of view is 60$^\circ$ and the image resolution is 1024$\times$768.
Finally, we map the 88 semantic classes from Replica to NYUv3-13, following~\cite{ZhiICCV2021,dai2017scannet}.
Our dataset consists of 12/3 scenes for train/test, respectively, resulting in 360/90 source-target pairs.
Note that this is 330$\times$ smaller than Hypersim. 
Yet, we show compelling results on Replica by pre-training \method on Hypersim.

\mypar{Results.}
Table~\ref{tab:replica_comp} reports the performance of our \method, trained on Hypersim, before fine-tuning (denoted as noft) and after fine-tuning, as well as a {\it Target View Oracle}.
The oracle fine-tunes the supervised semantic segmentation model on images from our Replica dataset and finds the optimal depth scale and shift for the test set.
We observe that \method's performance improves significantly when fine-tuning on Replica for both semantic segmentation and depth across all metrics. 
It is not surprising that fine-tuning on Replica improves performance as the scenes across the two datasets vary in both object appearance and geometry.
We also observe that performance is significantly higher than Hypersim (Table~\ref{tab:hypersim_comp}). 
Again, this is not a surprise, as Hypersim contains far more diverse and challenging scenes. 
However, the trends across the both datasets remain.

Fig.~\ref{fig:results_replica} shows predictions on Replica by \method. 
We show the 4 RGB inputs (left), the ground truth RGB , semantic and depth map for the novel target viewpoing (middle) and \method's predictions (right).
Fig.~\ref{fig:results_3d} shows 3D semantic reconstructions on two Replica test scenes.

\section{Conclusion}
\label{sec:conclusion}

We present \method, an end-to-end approach which learns a 3D representation of a scene from merely 4 input views. 
\method enables models to query its learnt 3D representation with a previously unseen target viewpoint of a novel sene to predict semantics and depth from that view, without access to any visual information (\eg RGB) from the view.
We discuss our work's limitations with an extensive quantitative and qualitative analysis.
We believe that we present a very promising direction to learn 3D from 2D data with lots of potential for exciting future work.

Regarding ethical risks, we do not foresee any immediate dangers. 
The datasets used in this work do not contain any humans or any other sensitive information.


\mypar{Acknowledgments}
We thank Shuaifeng Zhi for his help of Semantic-NeRF, Oleksandr Maksymets and Yili Zhao for their help of AI-Habitat, and Ang Cao, Chris Rockwell, Linyi Jin, Nilesh Kulkarni for helpful discussions.

{\small
\bibliographystyle{ieee_fullname}
\bibliography{local}
}

\clearpage
\appendix
\section{Qualitative Results}

\begin{figure*}[t]
  \centering
   \includegraphics[width=0.99\linewidth]{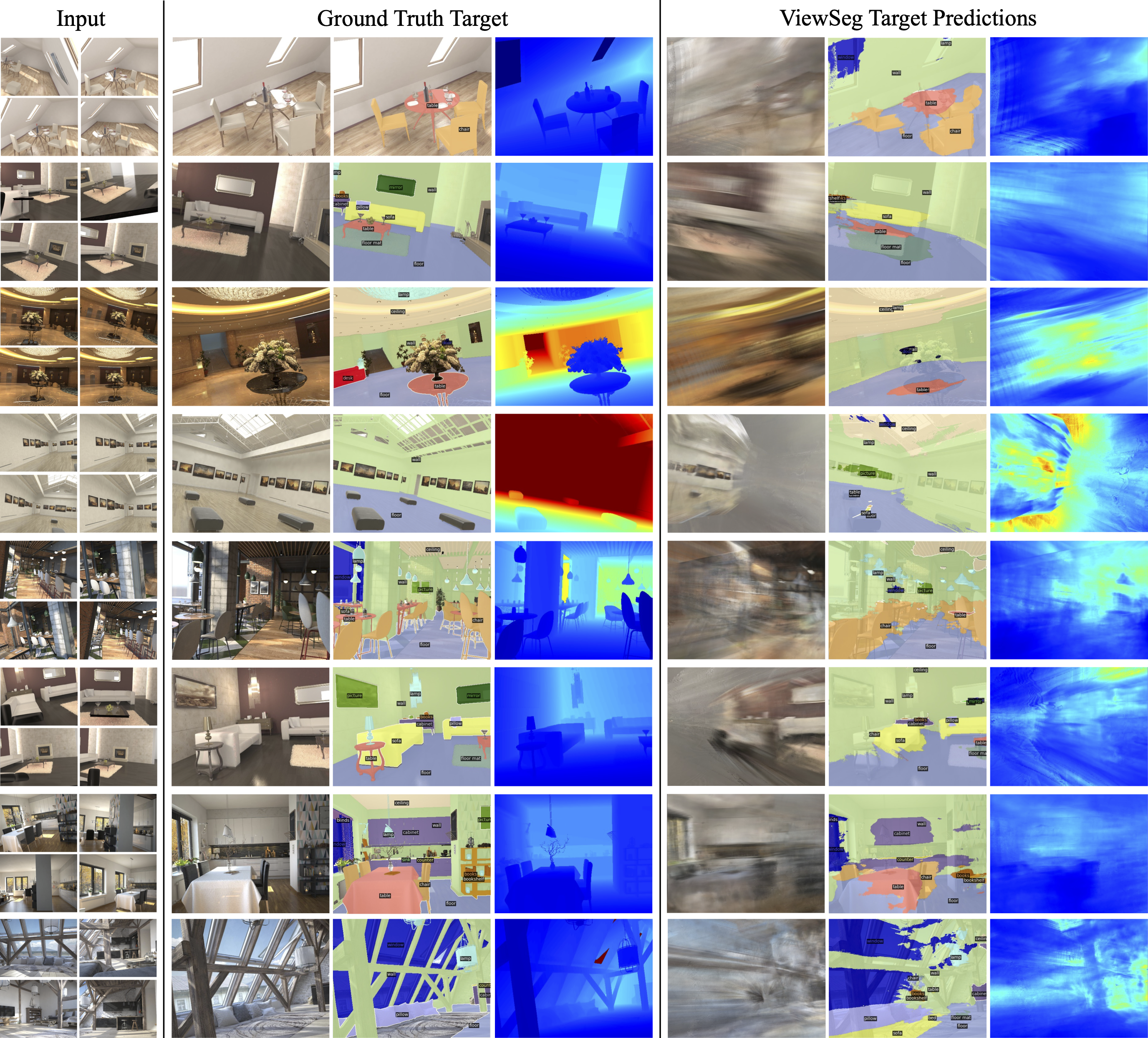}
   \vspace{-2mm}
   \caption{More predictions on Hypersim.
   For each example, we show the 4 input RGB views (left), the ground truth RGB, semantic and depth maps for the novel target view (middle) and \method's predictions (right).
   Our model does not have access to the true observations from the target view at test time. See our video animations.
   Depth: 0.1m \includegraphics[width=0.07\linewidth]{figures/depth_colormap.png} 20m.}
   \label{fig:hypersim_more}
   \vspace{-3mm}
\end{figure*}

\begin{figure*}[t]
  \centering
   \includegraphics[width=0.99\linewidth]{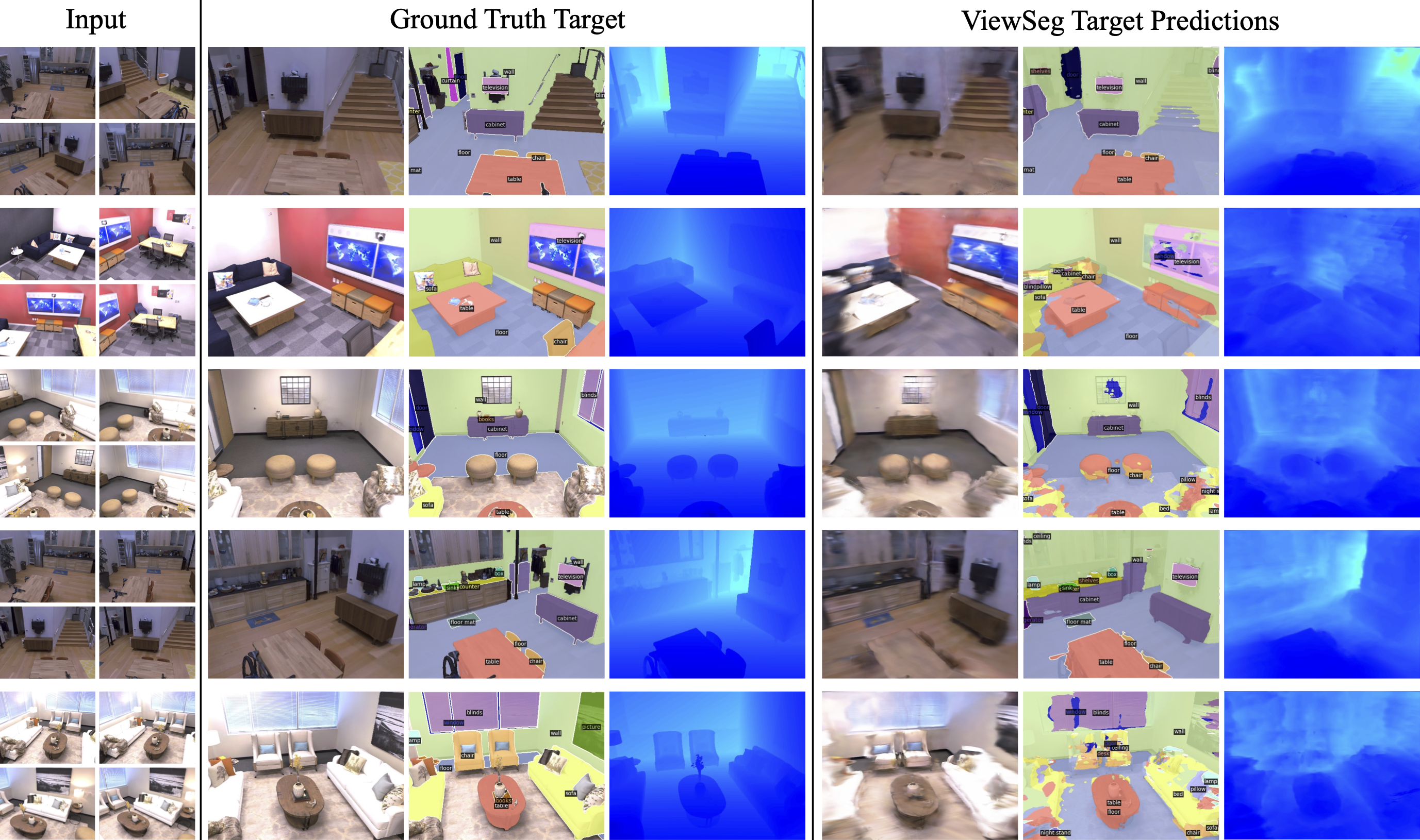}
   \caption{More predictions on Replica.
   For each example, we show the 4 input RGB views (left), the ground truth RGB, semantic and depth maps for the novel target view (middle) and \method's predictions (right).
   Our model does not have access to the true observations from the target view at test time. See our video animations.
   Depth: 0.1m \includegraphics[width=0.07\linewidth]{figures/depth_colormap.png} 20m.}
   \label{fig:replica_more}
   \vspace{-2mm}
\end{figure*}

Figure~\ref{fig:hypersim_more} shows more qualitative results on Hypersim. 
We draw a few interesting observations. 
Overall our model is able to segment objects and stuff from novel viewpoints and capture the underlying 3D geometry as seen in the predicted depth maps.
In addition, our model is able to generalize to completely unseen parts of a scene. 
For example, consider the 4$^\textrm{th}$ example in Figure~\ref{fig:hypersim_more} of a museum hall. 
Here, the four input images show the left side of the hall while the novel view queries the right side, which is not captured by the inputs. 
As seen from our RGB prediction, our model struggles to reconstruct that side of the scene in appearance space. 
It does much better in semantic space, by correctly placing the wall and the ceiling and by extending the floor. 
The predicted depth also shows that the model can reason about the geometry of the scene as the right wall starts close to the camera and extends backward consistent with the overall structure of the room.
Though both depth and semantic predictions are not perfect, they are evidence that our model has learnt to reason about 3D semantics.
We believe this is attributed to the scene priors our model has captured after being trained on hundreds of diverse scenes.
The 8$^\textrm{th}$ (last) example in Figure~\ref{fig:hypersim_more} leads to a similar conclusion. 
Our model correctly predicts the \emph{pillows} in the target viewpoint but additionally predicts a sofa which is sensibly placed relative to the location and extent of the predicted pillows. 
The sofa prediction, though not in the ground truth, is a reasonable one and is likely driven by the scene priors our model has captured during training; a line of pillows usually exists on a sofa.
Finally, the examples in  Figure~\ref{fig:hypersim_more} and Figure 3 in the main paper also reveal our work's limitations.
Our model does not make pixel-perfect predictions and often misses parts of objects or misplaces them by a few pixels in the target view.
It is likely that more training data would lead to significantly better predictions.
Figure~\ref{fig:replica_more} shows more qualitative results on Replica.

Last but not least, we provide video animations of our predictions in the supplementary folder. 
These complement the static visualizations in the pdf submissions and better demonstrate our model's ability to make predictions from novel viewpoints on novel scenes.

\section{Network Architecture}

\begin{figure*}[t]
  \centering
   \includegraphics[width=0.99\linewidth]{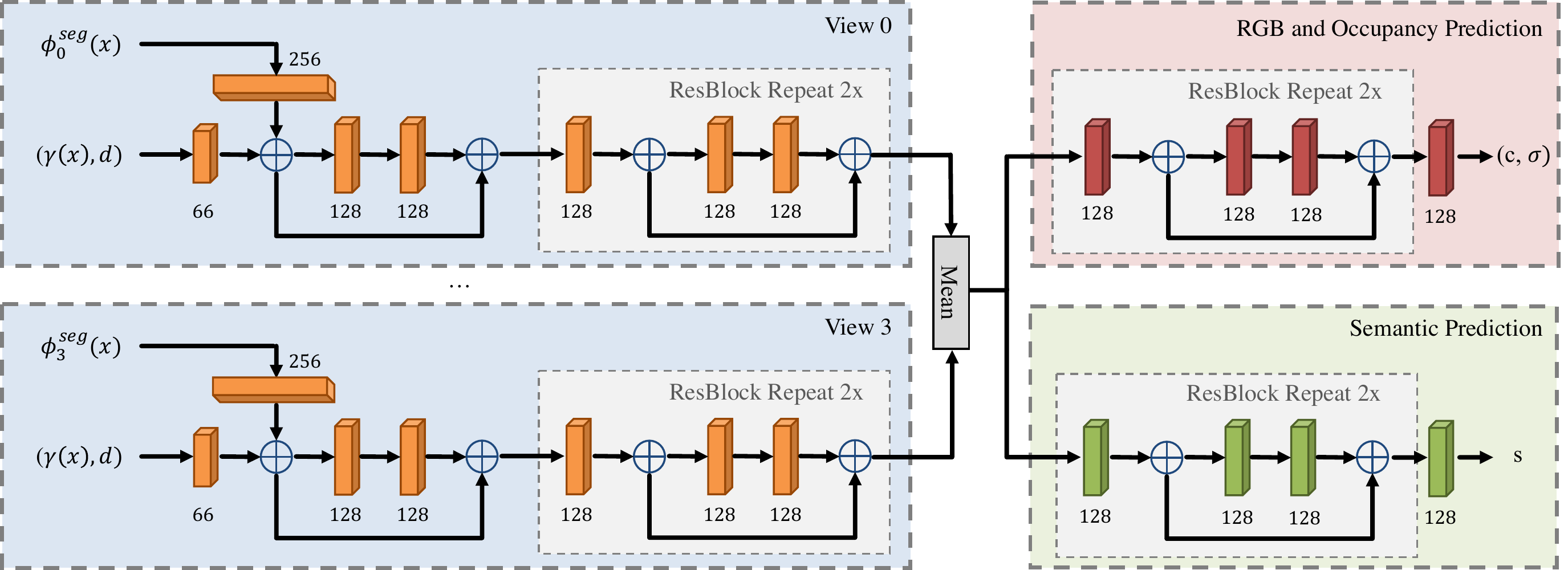}
   \caption{Detailed architecture of our semantic 3D representation $f$. Each cuboid is a linear layer where the number around it represents the input dimension. The output dimension is always 128 except the last layer of RGB and semantic prediction. Before the mean aggregation, the network takes inputs from each source view but weights are shared.}
   \label{fig:architecture_supp}
   \vspace{-2mm}
\end{figure*}

The detailed architecture of $f$ is illustrated in Figure \ref{fig:architecture_supp}. 
It takes as input a positional embedding of the 3D coordinates of $\mathbf{x}$, $\gamma(\mathbf{x})$, the viewing direction $\mathbf{d}$ and the semantic embeddings $\{\phi^\textrm{seg}_j\}_{j=1}^{N}$.
As output, $f$ produces
\begin{equation}
    (\cB, \, \sigma, \, \sB) = f\bigl( \gamma(\mathbf{x}), \, \mathbf{d}, \, \phi^\textrm{seg}_0(\mathbf{x}), \, ..., \, \phi^\textrm{seg}_{N-1}(\mathbf{x}) \bigr)
\end{equation}
where $\cB \in \mathbb{R}^3$ is the RGB color, $\sigma \in \mathbb{R}$ is the occupancy, and $\sB \in \mathbb{R}^{|\mathcal{C}|}$ is the distribution over semantic classes $\mathcal{C}$.
Hypersim~\cite{roberts2020hypersim} provides annotations for 40 semantic classes.
We discard \emph{\{otherstructure, otherfurniture, otherprop\}} and thus $|\mathcal{C}| = 37$ for both Hypersim~\cite{roberts2020hypersim} and Replica~\cite{straub2019replica}.

We largely follow PixelNeRF~\cite{yu2021pixelnerf} for the design of our network.
We deviate from PixelNeRF and use 128 instead of 512 for the dimension of hidden layers (as in NeRF~\cite{mildenhall2020nerf}) for a more compact network.
The dimension of the linear layer which inputs $\phi^\textrm{seg}_{j}(\mathbf{x})$ is set to 256 to match the dimension of semantic features from DeepLabv3+~\cite{chen2018encoder}.

\section{Training Details}

\mypar{Pretraining the Semantic Segmentation Module.}
We first pretrain DeepLabv3+~\cite{chen2018encoder} on ADE20k~\cite{zhou2019semantic}, which has 20,210 images for training and 2,000 images for validation.
We implement DeepLabv3+ in PyTorch with Detectron2~\cite{wu2019detectron2}.
We train on the ADE20k training set for 160k iterations with a batch size of 16 across 8 Tesla V100 GPUs. 
The model is initialized using ImageNet weights \cite{deng2009imagenet}.
We optimize with SGD and a learning rate of 1e-2. 
During training, we crop each input image to 512x512.
We remove the final layer, which predicts class probabilities, and use the output from the penultimate layer as our semantic encoder.
We do not freeze the model when training \method, allowing finetuning on Hypersim semantic categories.

\mypar{Training Details on Hypersim.}
We implement \method in PyTorch3D~\cite{ravi2020pytorch3d} and Detectron2~\cite{wu2019detectron2}.
We initialize the semantic segmentation module with ADE20k pretrained weights.
We train on the training set for 13 epochs with a batch size of 32 across 32 Tesla V100 GPUs.
The input and render resolution are set to 1024x768. 
We optimize with Adam and a learning rate of 5e-4. 
We follow the PixelNeRF~\cite{yu2021pixelnerf} strategy for ray sampling: We sample 64 points per ray in the coarse pass and 128 points per ray in the fine pass;
we sample 1024 rays per image.

\mypar{Training Details on Replica.}
We finetune our model on the Replica training set~\cite{straub2019replica}.
Replica has 18 scenes.
In practice, we find Replica does not have room-level annotations and our sampled source and target views can be at different rooms within the Replica apartments.
Hence, we exclude them from our data.
We split the rest 15 scenes into a train/val split of 12/3 scenes.
We use the same hyperparameters as Hypersim to finetune our model on Replica.

\section{Noisy Camera Experiment}

In our experiments, we assume camera poses both during training and evaluation.
We perform an additional ablation assuming \emph{noisy} cameras for both during training and testing. 
During evaluation, source view cameras are noisy but not the target camera, as we wish to compare to the target view ground truth.
We insert noise to the cameras by perturbing the rotation matrix with random angles sampled from $[-10^\circ, 10^\circ]$ in all three axis ($X$, $Y$ \& $Z$) .
This results in a significant camera noise and stretch tests our method under such conditions.

\begin{figure*}
  \centering
   \includegraphics[width=0.99\linewidth]{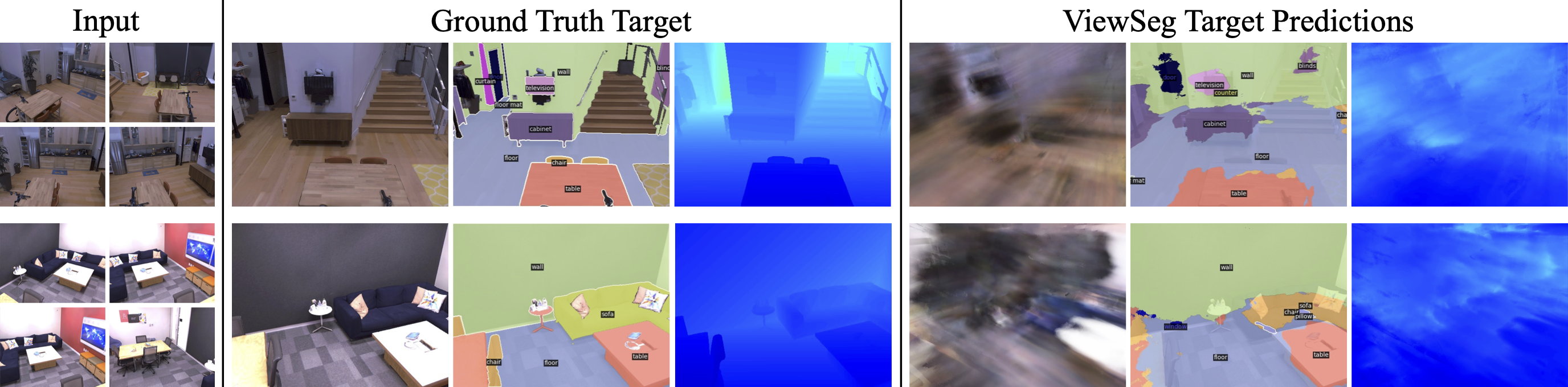}
   \caption{Predictions on Replica with noisy cameras.
   For each example, we show the 4 input RGB views (left), the ground truth RGB, semantic and depth maps for the novel target view (middle) and \method's predictions (right).
   Our model does not have access to the true observations from the target view at test time. Depth: 0.1m \includegraphics[width=0.07\linewidth]{figures/depth_colormap.png} 20m.}
   \label{fig:replica_noisy_camera}
   \vspace{-2mm}
\end{figure*}

Table~\ref{tab:replica_noisy_camera} shows results on the noisy Replica test set.
The 1$^\textrm{st}$ row shows the performance of our \method pre-trained on Hypersim and without any finetuning.
The 2$^\textrm{nd}$ row shows the performance of our model finetuned on the noisy Replica training set.
We observe that performance for both model variants is worse compared to having perfect cameras, as is expected.
Performance improves after training with camera noise, suggesting that our \method is able to generalize better when trained with noise in cameras. 
Figure~\ref{fig:replica_noisy_camera} shows qualitative results on the noisy Replica test set. 
The predicted RGB targets are significantly worse compared to the perfect camera scenario, which suggests that RGB prediction with imperfect cameras is very challenging.
However, the semantic predictions are much better computer to their RGB counterparts.
This shows that our approach is able to capture scene priors and generalize to new scenes even under imperfect conditions.

\begin{table*}
    \centering
    \scalebox{0.86}{
    \begin{tabular}{L{.13\linewidth}|S{.05\linewidth}S{.05\linewidth}S{.05\linewidth}S{.05\linewidth}S{.05\linewidth}S{.05\linewidth}|D{.05\linewidth}D{.07\linewidth}D{.07\linewidth}D{.07\linewidth}D{.08\linewidth}D{.09\linewidth}D{.09\linewidth}}
     Model & mIoU & IoU$^\textrm{T}$ & IoU$^\textrm{S}$ & fwIoU  & pACC & mACC  & L$_1$ ($\downarrow$) & Rel ($\downarrow$) & Rel$^\textrm{T}$ ($\downarrow$) & Rel$^\textrm{S}$ ($\downarrow$) & $\delta < 1.25$ & $\delta < 1.25^2$ \\
     \hhline
     \method noft & 8.57 & 35.2 & 61.3 & 40.3 & 18.3 & 57.9 & 1.10 & 0.253 & 0.240 & 0.268 & 0.579 & 0.840\\
     \method & 14.1 & 40.1 & 62.8 & 42.8 & 24.6 & 60.6 & 0.887 & 0.208 & 0.232 & 0.180 & 0.631 & 0.888\\
     \hline
   \end{tabular}
   } 
   \vspace{-2mm}
   \caption{Performance for semantic segmentation (\textcolor{llblue}{blue}) and depth (\textcolor{llgreen}{green}) on the Replica~\cite{straub2019replica} test set with noisy cameras.}
   \label{tab:replica_noisy_camera}
   \vspace{-3mm}
 \end{table*}

\section{Evaluation}

We report the complete set of depth metrics for all the tables in the main submission.
The comparison with PixelNeRF and CloudSeg is in Table~\ref{tab:supp_hypersim_comp}.
The ablation of different loss terms is in Table~\ref{tab:supp_hypersim_loss_ablations}.
The comparison of different backbones is in Table~\ref{tab:supp_hypersim_backbone_ablations}.
The input study is in Table~\ref{tab:supp_hypersim_views_ablations}. 
Our experiments on the Replica dataset are in Table~\ref{tab:supp_replica_comp}.

\begin{table*}[h!]
  \centering
  \scalebox{0.47}{
  \begin{tabular}{L{.1\linewidth}|S{.05\linewidth}S{.05\linewidth}S{.05\linewidth}S{.05\linewidth}S{.05\linewidth}S{.05\linewidth}|D{.05\linewidth}D{.06\linewidth}D{.07\linewidth}D{.07\linewidth}D{.07\linewidth}D{.07\linewidth}D{.08\linewidth}D{.09\linewidth}D{.09\linewidth}D{.09\linewidth}D{.1\linewidth}D{.1\linewidth}D{.09\linewidth}D{.1\linewidth}D{.1\linewidth}D{.09\linewidth}}
  Model & mIoU & IoU$^\textrm{T}$ & IoU$^\textrm{S}$ & fwIoU  & pACC & mACC  & L$_1$ ($\downarrow$) & L$_1^\textrm{T}$ ($\downarrow$) & L$_1^\textrm{S}$ ($\downarrow$) & Rel ($\downarrow$) & Rel$^\textrm{T}$ ($\downarrow$) & Rel$^\textrm{S}$ ($\downarrow$) & $\delta < 1.25$ & $\delta^\textrm{T} < 1.25$ & $\delta^\textrm{S} < 1.25$ & $\delta < 1.25^2$ & $\delta^\textrm{T} < 1.25^2$ & $\delta^\textrm{S} < 1.25^2$ & $\delta < 1.25^3$ & $\delta^\textrm{T} < 1.25^3$ & $\delta^\textrm{S} < 1.25^3$\\
 \hhline
 PixelNeRF++ & 1.58 & 14.9 & 47.9 & 17.9 & 3.63 & 35.9 & 2.80 & 2.69 & 2.90 & 0.746 & 0.856 & 0.653 & 0.300 & 0.276 & 0.319 & 0.531 & 0.500 & 0.557 & 0.689 & 0.663 & 0.712\\ 
 \cloud & 0.46 & 29.6 & 4.42 & 1.80 & 3.31 & 3.25 & 3.81 & 3.77 & 3.83 & 0.856 & 0.997 & 0.737 & 0.145 & 0.105 & 0.178 & 0.277 & 0.211 & 0.332 & 0.389 & 0.314 & 0.451 \\
 \method & \bf{17.1} & \bf{33.2} & \bf{58.9} &  \bf{44.8} & \bf{23.9} & \bf{62.2} & \bf{2.29} & \bf{2.18} & \bf{2.38} & \bf{0.646} & \bf{0.721} & \bf{0.584} & \bf{0.409} & \bf{0.393} & \bf{0.421} & \bf{0.656} & \bf{0.633} & \bf{0.676} & \bf{0.794} & \bf{0.772} & \bf{0.812} \\
 \hline
 Oracle & 40.0 & 58.1 & 71.3 & 66.6 & 52.1 & 79.1 & 0.96 & 1.10 & 0.83 & 0.235 & 0.317 & 0.163 & 0.731 & 0.651 & 0.800 & 0.898 & 0.848 & 0.942 & 0.954 & 0.925 & 0.978 \\ 
 \end{tabular}
 } 
 \vspace{-3mm}
 \caption{Extended version of Table 1 in the main paper. 
 }
 \label{tab:supp_hypersim_comp}
 \vspace{-3mm}
\end{table*}

\begin{table*}[h!]
  \centering
  \scalebox{0.45}{
  \begin{tabular}{L{.18\linewidth}|S{.05\linewidth}S{.05\linewidth}S{.05\linewidth}S{.05\linewidth}S{.05\linewidth}S{.05\linewidth}|D{.05\linewidth}D{.06\linewidth}D{.07\linewidth}D{.07\linewidth}D{.07\linewidth}D{.07\linewidth}D{.08\linewidth}D{.09\linewidth}D{.09\linewidth}D{.09\linewidth}D{.1\linewidth}D{.1\linewidth}D{.09\linewidth}D{.1\linewidth}D{.1\linewidth}D{.09\linewidth}}
    \method loss & mIoU & IoU$^\textrm{T}$ & IoU$^\textrm{S}$ & fwIoU  & pACC & mACC  & L$_1$ ($\downarrow$) & L$_1^\textrm{T}$ ($\downarrow$) & L$_1^\textrm{S}$ ($\downarrow$) & Rel ($\downarrow$) & Rel$^\textrm{T}$ ($\downarrow$) & Rel$^\textrm{S}$ ($\downarrow$) & $\delta < 1.25$ & $\delta^\textrm{T} < 1.25$ & $\delta^\textrm{S} < 1.25$ & $\delta < 1.25^2$ & $\delta^\textrm{T} < 1.25^2$ & $\delta^\textrm{S} < 1.25^2$ & $\delta < 1.25^3$ & $\delta^\textrm{T} < 1.25^3$ & $\delta^\textrm{S} < 1.25^3$\\
 \hhline
 \emph{w/o} photometric loss  & 16.9 & 30.8 & 58.7 & 44.8 & 22.7 & 62.5 & 2.49 & 2.35 & 2.61 & 0.677 & 0.750 & 0.615 & 0.359 & 0.347 & 0.363 & 0.611 & 0.594 & 0.625 & 0.764 & 0.744 & 0.780\\
 \emph{w/o} semantic loss & - & - & - & - & - & - & 2.58 & 2.52 & 2.63 & 0.787 & 0.919 & 0.678 & 0.345 & 0.317 & 0.369 & 0.587 & 0.548 & 0.621 & 0.740 & 0.704 & 0.770\\
 \emph{w/o} source view loss       & 14.3 & 28.2 & 57.9 & 28.2 & 19.3 & 61.1 & 2.37 & 2.27 & 2.45 & 0.683 & 0.764 & 0.615 & 0.397 & 0.378 & 0.413 & 0.649 & 0.623 & 0.670 & 0.785 & 0.761 & 0.806\\
   \emph{w/o} viewing dir       & 16.0 & 33.1 & \bf{59.2} & \bf{44.9} & 21.5 & 62.1 & 2.53 & 2.38 & 2.65 & 0.708 & 0.783 & 0.646 & 0.354 & 0.351 & 0.356 & 0.602 & 0.593 & 0.610 & 0.759 & 0.744 & 0.772 \\
   final                        & \bf{17.1} & \bf{33.2} & 58.9 & 44.8 & \bf{23.9} & \bf{62.2} & \bf{2.29} & \bf{2.18} & \bf{2.38} & \bf{0.646} & \bf{0.721} & \bf{0.584} & \bf{0.409} & \bf{0.393} & \bf{0.421} & \bf{0.656} & \bf{0.633} & \bf{0.676} & \bf{0.794} & \bf{0.772} & \bf{0.812} \\
 \end{tabular}
 } 
 \vspace{-3mm}
 \caption{Extended version of Table 2 in the main paper. 
 }
 \label{tab:supp_hypersim_loss_ablations}
 \vspace{-2mm}
\end{table*}

\begin{table*}[h!]
  \centering
 \scalebox{0.44}{
  \begin{tabular}{L{.22\linewidth}|S{.05\linewidth}S{.05\linewidth}S{.05\linewidth}S{.05\linewidth}S{.05\linewidth}S{.05\linewidth}|D{.05\linewidth}D{.06\linewidth}D{.07\linewidth}D{.07\linewidth}D{.07\linewidth}D{.07\linewidth}D{.08\linewidth}D{.09\linewidth}D{.09\linewidth}D{.09\linewidth}D{.1\linewidth}D{.1\linewidth}D{.09\linewidth}D{.1\linewidth}D{.1\linewidth}D{.09\linewidth}}
    \method backbone & mIoU & IoU$^\textrm{T}$ & IoU$^\textrm{S}$ & fwIoU  & pACC & mACC  & L$_1$ ($\downarrow$) & L$_1^\textrm{T}$ ($\downarrow$) & L$_1^\textrm{S}$ ($\downarrow$) & Rel ($\downarrow$) & Rel$^\textrm{T}$ ($\downarrow$) & Rel$^\textrm{S}$ ($\downarrow$) & $\delta < 1.25$ & $\delta^\textrm{T} < 1.25$ & $\delta^\textrm{S} < 1.25$ & $\delta < 1.25^2$ & $\delta^\textrm{T} < 1.25^2$ & $\delta^\textrm{S} < 1.25^2$ & $\delta < 1.25^3$ & $\delta^\textrm{T} < 1.25^3$ & $\delta^\textrm{S} < 1.25^3$\\
 \hhline
 DLv3+~\cite{chen2018encoder} + ADE20k~\cite{zhou2019semantic} & \bf{17.1} & \bf{33.2} & 58.9 & 44.8 & \bf{23.9} & 62.2 & 2.29 & 2.18 & 2.38 & 0.645 & 0.721 & 0.584 & 0.409 & 0.393 & 0.421 & 0.656 & 0.633 & 0.676 & 0.794 & 0.772 & 0.812 \\
   DLv3+~\cite{chen2018encoder} + IN~\cite{deng2009imagenet} & 16.3 & \bf{33.2} & \bf{59.2} & \bf{45.2} & 22.0 & \bf{62.5} & \bf{2.28} & \bf{2.17} & \bf{2.36} & \bf{0.614} & \bf{0.682} & \bf{0.559} & \bf{0.415} & \bf{0.400} & \bf{0.427} & \bf{0.663} & \bf{0.640} & \bf{0.682} & \bf{0.799} & \bf{0.776} & \bf{0.818} \\
   ResNet34~\cite{he2016deep} + IN~\cite{deng2009imagenet} & 7.45 & 21.7 & 55.9 & 37.1 & 11.2 & 56.1 & 2.67 & 2.51 & 2.81 & 0.712 & 0.815 & 0.626 & 0.320 & 0.304 & 0.333 & 0.562 & 0.541 & 0.580 & 0.720 & 0.702 & 0.736\\
 \end{tabular}
 } 
 \vspace{-2mm}
 \caption{Extended version of Table 3 in the main paper. 
 }
 \label{tab:supp_hypersim_backbone_ablations}
 \vspace{-4mm}
\end{table*}

\begin{table*}
  \centering
  \scalebox{0.47}{
  \begin{tabular}{L{.1\linewidth}|S{.05\linewidth}S{.05\linewidth}S{.05\linewidth}S{.05\linewidth}S{.05\linewidth}S{.05\linewidth}|D{.05\linewidth}D{.06\linewidth}D{.07\linewidth}D{.07\linewidth}D{.07\linewidth}D{.07\linewidth}D{.08\linewidth}D{.09\linewidth}D{.09\linewidth}D{.09\linewidth}D{.1\linewidth}D{.1\linewidth}D{.09\linewidth}D{.1\linewidth}D{.1\linewidth}D{.09\linewidth}}
 \method & mIoU & IoU$^\textrm{T}$ & IoU$^\textrm{S}$ & fwIoU  & pACC & mACC  & L$_1$ ($\downarrow$) & L$_1^\textrm{T}$ ($\downarrow$) & L$_1^\textrm{S}$ ($\downarrow$) & Rel ($\downarrow$) & Rel$^\textrm{T}$ ($\downarrow$) & Rel$^\textrm{S}$ ($\downarrow$) & $\delta < 1.25$ & $\delta^\textrm{T} < 1.25$ & $\delta^\textrm{S} < 1.25$ & $\delta < 1.25^2$ & $\delta^\textrm{T} < 1.25^2$ & $\delta^\textrm{S} < 1.25^2$ & $\delta < 1.25^3$ & $\delta^\textrm{T} < 1.25^3$ & $\delta^\textrm{S} < 1.25^3$\\
 \hhline
 \emph{w/} 4 views & \bf{17.1} & \bf{33.2} & \bf{58.9} & \bf{44.8} & \bf{23.9} & \bf{62.2} & \bf{2.29} & \bf{2.18} & \bf{2.38} & \bf{0.646} & \bf{0.721} & \bf{0.584} & \bf{0.408} & \bf{0.393} & \bf{0.421} & \bf{0.656} & \bf{0.633} & \bf{0.676} & \bf{0.794} & \bf{0.772} & \bf{0.812}\\
 \emph{w/} 3 views & 15.5 & 31.3 & 58.7 & 43.9 & 20.8 & 61.5 & 2.39 & 2.25 & 2.49 & 0.652 & 0.730 & 0.587 & 0.387 & 0.376 & 0.395 & 0.634 & 0.617 & 0.648 & 0.777 & 0.759 & 0.793\\ 
 \emph{w/} 2 views & 13.6 & 27.4 & 57.7 & 41.9 & 18.2 & 60.2 & 2.57 & 2.49 & 2.64 & 0.765 & 0.878 & 0.672 & 0.363 & 0.339 & 0.383 & 0.605 & 0.574 & 0.633 & 0.751 & 0.721 & 0.776\\ 
 \emph{w/} 1 view  & 11.6 & 24.9 & 56.5 & 39.7 & 15.8 & 57.9 & 2.62 & 2.52 & 2.70 & 0.734 & 0.828 & 0.657 & 0.332 & 0.322 & 0.339 & 0.562 & 0.541 & 0.580 & 0.710 & 0.686 & 0.730\\
 \end{tabular}
 } 
 \vspace{-2mm}
 \caption{Extended version of Table 4 in the main paper.}
 \label{tab:supp_hypersim_views_ablations}
\end{table*}

\begin{table*}
  \centering
  \scalebox{0.46}{
  \begin{tabular}{L{.15\linewidth}|S{.05\linewidth}S{.05\linewidth}S{.05\linewidth}S{.05\linewidth}S{.05\linewidth}S{.05\linewidth}|D{.05\linewidth}D{.06\linewidth}D{.07\linewidth}D{.07\linewidth}D{.07\linewidth}D{.07\linewidth}D{.08\linewidth}D{.09\linewidth}D{.09\linewidth}D{.09\linewidth}D{.1\linewidth}D{.1\linewidth}D{.09\linewidth}D{.1\linewidth}D{.1\linewidth}D{.09\linewidth}}
 \method & mIoU & IoU$^\textrm{T}$ & IoU$^\textrm{S}$ & fwIoU  & pACC & mACC  & L$_1$ ($\downarrow$) & L$_1^\textrm{T}$ ($\downarrow$) & L$_1^\textrm{S}$ ($\downarrow$) & Rel ($\downarrow$) & Rel$^\textrm{T}$ ($\downarrow$) & Rel$^\textrm{S}$ ($\downarrow$) & $\delta < 1.25$ & $\delta^\textrm{T} < 1.25$ & $\delta^\textrm{S} < 1.25$ & $\delta < 1.25^2$ & $\delta^\textrm{T} < 1.25^2$ & $\delta^\textrm{S} < 1.25^2$ & $\delta < 1.25^3$ & $\delta^\textrm{T} < 1.25^3$ & $\delta^\textrm{S} < 1.25^3$\\
 \hhline
 \method noft & 13.2 & 44.8 & 56.0 & 51.4 & 27.1 & 66.8 & 0.982 & 0.851 & 1.138 & 0.222 & 0.194 & 0.254 & 0.623 & 0.687 & 0.546 & 0.880 & 0.905 & 0.850 & 0.968 & 0.974 & 0.961 \\
 \method & 30.2 & 56.2 & 62.8 & 62.3 & 48.4 & 75.6 & 0.550 & 0.510 & 0.597 & 0.130 & 0.130 & 0.130 & 0.851 & 0.857 & 0.844 & 0.961 & 0.953 & 0.972 & 0.986 & 0.980 & 0.992\\
 \hline
 Oracle & 56.2 & 76.8 & 78.0 & 90.1 & 79.4 & 93.8 & 0.226 & 0.230 & 0.220 & 0.058 & 0.065 & 0.050 & 0.976 & 0.965 & 0.991 & 0.998 & 0.996 & 0.999 & 1.000 & 1.000 & 1.000\\
 \end{tabular}
 } 
 \vspace{-2mm}
 \caption{Extended version of Table 5 in the main paper. 
 }
 \label{tab:supp_replica_comp}
 \vspace{-3mm}
\end{table*}



\end{document}